\documentclass[journal]{IEEEtran}
\usepackage{amsmath,amsfonts,mathrsfs,amssymb,amsthm}
\usepackage[ruled]{algorithm2e} % add "vlined" to add End, “longend" to End as endif, endfor
\SetKwRepeat{Do}{do}{while}
\usepackage{multirow}
\usepackage[table,xcdraw]{xcolor}
\usepackage[caption=false,font=normalsize,labelfont=sf,textfont=sf]{subfig}
\usepackage{textcomp}
\usepackage{paralist}
\usepackage{stfloats}
\usepackage{xurl}
\usepackage{verbatim}
\usepackage{graphicx}
\DeclareMathAlphabet{\mathcal}{OMS}{cmsy}{m}{n}
\usepackage{lipsum}
\usepackage{cuted} % one column package
\usepackage{orcidlink}
\usepackage{booktabs}

\usepackage{hyperref}
\hypersetup{colorlinks=true,
	linkcolor=blue,
	anchorcolor=blue,
	citecolor=blue}
\usepackage{cite}

\newtheorem{assumption}{Assumption}
\newtheorem{remark}{Remark}
\newtheorem{theorem}{Theorem}

\newcommand{\mb}[1]{\mathbf{#1}} 

\newcommand{\mr}[1]{\mathrm{#1}} 
\newcommand{\mc}[1]{\mathcal{#1}}
\newcommand{\hb}[1]{\hat{\mathbf{#1}}}
\newcommand{\pt}{\partial}
\newcommand{\mrd}{\mathrm{d}}
\newcommand{\sgn}[1]{\mbox{sgn}(#1)}

\newcommand{\figref}[1]{Fig. \ref{#1}}

\def\BibTeX{{\rm B\kern-.05em{\sc i\kern-.025em b}\kern-.08em
    T\kern-.1667em\lower.7ex\hbox{E}\kern-.125emX}}
\usepackage{balance}
\begin{document}
\title{Closed-Loop Magnetic Control of Medical Soft Continuum 
Robots for Deflection}
\author{Zhiwei~Wu$^{\orcidlink{https://orcid.org/0000-0002-3957-3063}}$,~\IEEEmembership{Graduate~Student~Member}, 
Jinhui Zhang$^{\orcidlink{https://orcid.org/0000-0002-2405-894X}}$
\thanks{
Z. Wu, J. Zhang are with the School of Automation, Beijing Institute of Technology, Beijing, 100081, China (e-mail: 3120220885@bit.edu.cn; zhangjinh@bit.edu.cn). \textit{(Corresponding author: Jinhui Zhang, phone: +86 10 68914350; fax: +86 10 68914382)}.
}}

% \markboth{Journal of \LaTeX\ Class Files,~Vol.~18, No.~9, September~2020}%
% {How to Use the IEEEtran \LaTeX \ Templates}

\maketitle

\begin{abstract}
Magnetic soft continuum robots (MSCRs) have emerged as powerful devices in endovascular interventions owing to their hyperelastic fibre matrix and enhanced magnetic manipulability. Effective closed-loop control of tethered magnetic devices contributes to the achievement of autonomous vascular robotic surgery. In this article, we employ a magnetic actuation system equipped with a single rotatable permanent magnet to achieve closed-loop deflection control of the MSCR. To this end, we establish a differential kinematic model of MSCRs exposed to non-uniform magnetic fields. The relationship between the existence and uniqueness of Jacobian and the geometric position between robots is deduced. The control direction induced by Jacobian is demonstrated to be crucial in simulations. Then, the corresponding quasi-static control (QSC) framework integrates a linear extended state observer to estimate model uncertainties. Finally, the effectiveness of the proposed QSC framework is validated through comparative trajectory tracking experiments with the PD controller under external disturbances. Further extensions are made for the Jacobian to path-following control at the distal end position. The proposed control framework prevents the actuator from reaching the joint limit and achieves fast and low error-tracking performance without overshooting.
% 所提出的控制框架能够有效避免执行器达到关节极限，并实现了无超调的快速低误差跟踪性能。
% 我们

% 磁软连续体机器人成为血管介入手术提供新的强有力工具，由于其亚毫米级的尺寸, 超弹性纤维基质，和增强的磁可操纵性。这里，我们使用装备了单一可旋转永磁铁的磁驱动系统对MSCR完成闭环偏转控制。为此，建立了暴露在非匀强磁场中MSCR的微分运动学模型。通过分析旋转磁铁到MSCR的工作空间映射，讨论了零空间的存在性，并添加装置软限位回避雅可比奇异值。提出的准静态控制框架能够估计模型的不确定性。最后通过和PD控制器在包含外部扰动的对比实验证实了提出的QSC框架的有效性。

% 
% The feedback loop is accomplished by a visual servoing platform.

% Abstract 
\end{abstract}


\section{Introduction}
\IEEEPARstart{M}{agnetically-actuated} robotic-assisted surgery has garnered considerable attention and found extensive applications in vascular interventions recently. A well-established magnetic-actuated surgical robotic system typically consists of a designable and programmable magnetic field generator, a magnetic continuum robot capable of precise manipulation within the patient's vasculature using the generated field, and various complementary devices. In comparison to conventional robot-assisted interventions, the magnetic field demonstrates remarkable and secure penetration through human tissues, facilitating contactless object manipulation \cite{schenckSafetyStrongStatic2000}. The inherent compliance and flexibility of the magnetic continuum robot enable predictable deflection without applying torsion, thereby allowing convenient navigation through intricate and narrow vasculatures, including bifurcations.  

Systems equipped with electromagnets have been developed as laboratorial and commercial products for remote navigation of magnetic devices \cite{kongIntegratedLocomotionDeformation2021, leeSteeringTunnelingStent2021, fischerUsingMagneticFields2022}. The systems are designed to precisely manipulate the directionality of the generated magnetic field, simplifying the control challenges associated with magnetic devices. However, the structural openness of the electromagnetic systems restricts the size of the workspace, and the resulting thermal effects may pose potential safety risks to the human body \cite{hwang2020review}. In contrast, permanent magnets serve as ideal actuating devices in clinical settings, providing stable and powerful magnetic fields. Cutting-edge commercial systems incorporate a pair of large permanent magnets symmetrically positioned on either side of the patient \cite{carpi2009stereotaxis}, obtaining FDA certification and demonstrating significant success in cardiac ablation procedures \cite{StereotaxisEarnsFDA2020}. However, the expensive price, considerable volume, and challenges in downsizing the equipment pose obstacles to its deployment in primary healthcare facilities. Alternatively, the single permanent magnet offers affordability and compact size, allowing for convenient attachment with robot arms and enabling remote manipulation \cite{mahoneyFivedegreeoffreedomManipulationUntethered2016}. This compact actuating system, initially used in endoscopy departments, has been recognized for its feasibility in neurovascular interventions \cite{kimTeleroboticNeurovascularInterventions2022}. The achievement is mostly attributed to the development of novel magnetic soft continuum robots (MSCRs) fabricated using hard-magnetic soft composites \cite{kimFerromagneticSoftContinuum2019}. MSCRs evade the intravascular detachment of micromagnets in commercially available magnetic devices while also providing enhanced slimness and flexibility beyond the magnet thickness limit.

In advancing towards high-level/full autonomy, closed-loop control of tethered magnetic devices is a critical component in endovascular robotic surgery \cite{haideggerAutonomySurgicalRobots2019}. It effectively mitigates the complex nonlinear behaviour of continuum robots and assists clinical practitioners in shifting their focus from operational tasks to more intricate medical procedures. The electromagnetic system has emerged as the initial choice for implementing closed-loop control due to its capability of generating simple and uniform magnetic fields \cite{edelmannMagneticControlContinuum2017,thomasDesignSensingControl2022}. Although attempts have been made to incorporate permanent magnet systems with multi-actuate-magnets \cite{pittiglioClosedLoopStatic2023}, the prevalent approach still relies on open-loop control schemes \cite{abolfathiIndependentHybridMagnetic2023, liuIterativeJacobianBasedInverse2017,linPositionOrientationControl2023, 10381486}, possibly because of the complex geometric shape of the dipole field generated by permanent magnets. Therefore, the aim of this paper is to establish a differential kinematic model for magnetic soft composite material robots in non-uniform gradient magnetic fields and propose an observer-integrated quasi-static controller to achieve closed-loop deflection control with only a single actuating magnet. The main contributions can be summarized into the following threefolds:
\begin{enumerate}
    \item The three-dimensional kinematic model is established for the MSCR under a magnetic dipole field. Theorems to determine the minimum distance between the robot and the magnet are proposed, ensuring the existence and uniqueness of the Jacobian. The model accuracy is verified by comparing the singularities with experimental measurements.

    \item The controllability of the differential kinematic model for MSCR is proved. Based on this model, a quasi-static controller (QSC) for closed-loop deflection control is designed, incorporating a linear extended state observer to compensate for model uncertainties and disturbances.

    \item A visual detection algorithm for tracking the tip deflection angle of MSCR is deployed. The control performance of QSC is compared with that of a PD controller under various tracking signals and external disturbances. The Jacobian-based to distal end positional control is further explored.
\end{enumerate}

% 以下三个方面：
% 1.方程和解存在唯一的必要条件
% 2.控制器算一个
% 3.视觉平台和实验，Jacobian在位置控制上的应用

The remainder of this paper is organized as follows: Section \ref{sec:mathematicalModeling} studies the mathematical model of the permanent magnet and MSCR. Section \ref{sec:quasi-staticcontrol} discusses the Jacobian singularity and proposes the quasi-static controller. Simulations and experiments are conducted in Section \ref{sec:experiment}. Section \ref{sec:extension} explores the position control of the MCSR, and \ref{sec:conclusions} concludes this paper.
% 带有电磁铁的系统已经被开发为实验室级别/商业的产品用于磁设备的远程导航(CGCI, CardioMag)。电磁系统可以设计精确操纵产生的磁场的方向性，从而简化磁设备的控制难度。然而，电磁系统的结构开放性较低而限制了工作空间的大小，并且电磁铁热效应可能对人体产生潜在的安全风险。相比之下，永磁铁作为临床上理想的驱动设备能够提供稳定、强大的的磁场。前沿的商业化系统包含一对在病人两侧对称摆放的巨大永磁铁，已取得FDA认证并完成了大量心脏消融手术。然而，昂贵的造价和在设备小型化上存在困难令其在向基层医院部署时面临挑战。单一永磁铁价格低廉，外形小巧，可以很方便地固定在机械臂末端并实现远程操控。这一率先在内窥镜科室使用的结构紧凑的驱动系统被发掘了在神经血管介入的可行性。这一成就得益于利用hard-magnetic soft composites制作的novel磁驱动机器人，被称为magnetic soft continuum robots (MSCRs). 这一机器人可以规避市售磁性导丝/导管的micromagnets脱落的麻烦 \cite{Ferro}，同时更细和灵活，超越了磁铁厚度限制。

% Endovascular robotic surgery在迈向high-level/full autonomy的过程中，约束磁性设备的闭环控制是关键的一环 \cite{auto}。它能够掩盖连续体机器人的复杂非线性行为，协助临床医生将注意力从操作任务移到更复杂的医疗任务中。电磁系统成为了闭环控制实施的首先设备，由于其产生的简单，均匀的磁场。尽管一些在包含多个永磁铁系统上的尝试取得效果，单磁铁系统普遍采用的仍然是开环控制方案，这可能是受到永磁铁产生的偶极场的复杂几何形状的阻碍。

% 在这篇文章，我们仅使用一个附着在机械臂末端的永磁铁实现了对MSCR的闭环偏转控制。文章的主要贡献可以被总结为下面的三个部分：
% summarized as the following threefolds:
% 1. 建立了一套基于视觉的磁驱动系统，包含结构改进的MSCR和小型可旋转磁性装置。

\section{Mathematical Modeling}
\label{sec:mathematicalModeling}
Throughout the technical note, vectors are denoted in lower case, bold font (e.g. $\mb{x}$), and with an extra `hat' symbol to be unit (e.g. $\hat{\mb{x}}$). Scalars are denoted by Italy font (e.g. $\mathit{X}$), and matrices are denoted by Roman font (e.g. $\mr{X}$). $\|\cdot\|$ denotes the $\mathcal{L}_2$ norm for any tensor. The identity matrix is denoted by $\mathbb{I}$ with a subscript indicating the appropriate dimensions.

\subsection{Small Rotatable Magnetic Equipment}
\label{sec:II-A}
\begin{figure*}[t]
	\centering
	\includegraphics[width=\textwidth]{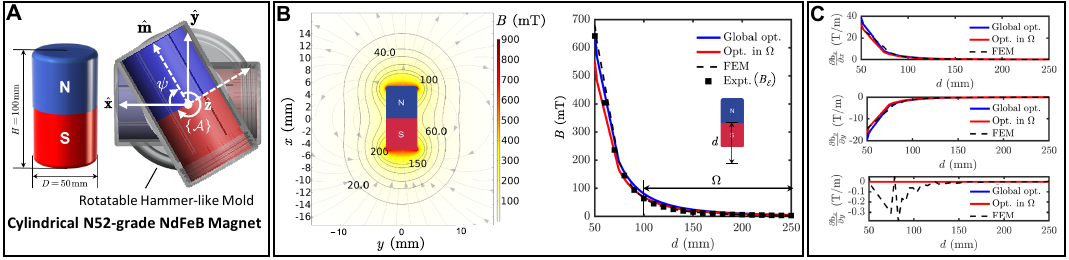}
	\caption{ Small rotatable magnetic equipment. (\textbf{A}) Cylindrical N52-grade NdFeB magnet loaded in a rotatable hammer-like mold. (\textbf{B}) (left) Axis-symmetric magnetic field distribution around the magnet with remanence $B_{re}=1.44$ $\mr{T}$. (right) Comparison of the analytical solutions of the magnetic flux density $\|\mb{b}\|$ (noted in $B$) along the principle axis at distance $d$ with the finite element method (FEM) and experiments. (\textbf{C}) Comparison of the spatial gradient magnetic field along the principle axis of the magnet in the distance $d$ between analytical solutions and FEM results ($\hat{\mb{x}}_{\mc{A}}$-$\hat{\mb{y}}_{\mc{A}}$ plane).}
 % Small rotatable magnetic equipment. (a) Cylindrical N52-grade NdFeB magnet loaded in a rotatable hammer-like mold. The clockwise rotation angle along the magnet's center is denoted in $\psi$. (b) (left) Axis-symmetric magnetic field distribution around the magnet with remanence $B_{re}=1.44$ $\mr{T}$. The gray lines with arrows indicate field directions and the black lines illustrate the contour of the magnetic flux density. (right) Comparison of the analytical solutions of the magnetic flux density $\|\mb{b}\|$ (solid lines, noted in $B$) along the principle axis at distance $d$ with the finite element method (FEM) and experiments. The parameter of the analytical model is optimally estimated by the least squares fitting for the global (blue line) and within the working distance $\Omega$ (red line) optimally. (c) Comparison of the spatial gradient magnetic field along the principle axis of the magnet in the distance $d$ between analytical solutions and FEM results ($\hat{\mb{x}}_{\mc{A}}$-$\hat{\mb{y}}_{\mc{A}}$ plane).}
	\label{fig:MagnetAnalysis}
\end{figure*}
%小型可旋转磁设备。(A)N52-级钕铁硼圆柱形磁铁装载在可旋转锤形模具中。

% 永磁铁作为磁性设备的良好驱动器，其产生的非均匀磁场模型已得到广泛研究。将磁铁嵌入设计的锤形模具以成为机械臂的末端执行器，可以通过改变磁铁的空间位置和方向引起目标位置的磁场变化，从而驱动磁性设备。被称为小型可旋转磁性设备的末端机构包含...
Permanent magnets, serving as effective drivers for magnetic devices, are capable of generating non-uniform gradient magnetic fields. Embedding magnets within a designed hammer-shaped mold to attach to a robotic arm allows for the manipulation of magnetic devices by altering the spatial positioning and orientation of the magnet. The end-effector mechanism, referred to as the small rotatable magnetic equipment (RME), contains an axially magnetized cylindrical N52-grade NdFeB magnet with diameter $D=50$ $\mr{mm}$, height $H=100$ $\mr{mm}$, and remanence $B_{re}=1.44$ $\mr{T}$. The magnet is loaded in a rotatable hammer-like mold with an attached frame $\{\mc{A}\}$ positioned at the geometric center $\mb{p}_{\mc{A}}\in\mathbb{R}^3$, where the $\hat{\mb{x}}_{\mc{A}}$-axis initially coincides with the dipole moment $\mb{m}_{\mc{A}}\in\mathbb{R}^{3}$ (the vector from south to north pole). As shown in \figref{fig:MagnetAnalysis}.\textbf{A}, the magnet can rotate along the $\hat{\mb{z}}_{\mc{A}}$-axis that is perpendicular to the principal axis. The rotation angle $\psi$ is defined to be clockwise positive, yielding the relations between $\psi$ and $\hat{\mb{m}}_{\mc{A}}$ as $\hat{\mb{m}}_{\mc{A}}=[\cos\psi,\sin\psi,0]^\top$. The magnetic field $\mb{b}(\mb{p},\hat{\mb{m}}_{\mc{A}})\in\mathbb{R}^3$ generated by the cylindrical magnet in any spatial point $\mb{p}_{\mc{S}}\in\mathbb{R}^3$ is assumed to be modeled by the point dipole model 
\cite{petruskaOptimalPermanentMagnetGeometries2013}, which is provided by
\begin{equation}
	\label{eqn:magneticfield}
	\mb{b}(\mb{p},\hat{\mb{m}}_{\mc{A}}) = \frac{\mu_0M_{\mc{A}}}{4\pi\|\mb{p}\|^3}\left(3\hat{\mb{p}}\hat{\mb{p}}^\top-\mathbb{I}_3\right)\hat{\mb{m}}_{\mc{A}},
\end{equation}
where $\mb{p}=\mb{p}_{\mc{S}}-\mb{p}_{\mc{A}}$, $\mu_0$ denotes the space permeability, and $M_{\mc{A}}=\|\mb{m}_{\mc{A}}\|$ is the dipole moment modulus to be estimated. The magnetic field distribution is simulated using the finite-element method (FEM) via COMSOL v5.6 and demonstrated in \figref{fig:MagnetAnalysis}.\textbf{B} (left). The field directions, indicated by gray lines with arrows, show that the S pole of the magnet performs attraction to magnetic devices while the N pole performs repulsion. Manipulation distance from $\mb{p}_{\mc{A}}$ to magnetic devices is limited to $\Omega=[100,250]$ $(\mr{mm})$ for surgical safety concerns. Experimental results of the magnetic field modulus $B$ in the distance $d$ along the principal axis of the magnet are plotted against the FEM results in \figref{fig:MagnetAnalysis}.\textbf{B} (right). The magnetic dipole moment modulus $M_{\mc{A}}$ is estimated for global and within the working distance $\Omega$ optimally using the least square fitting (LSF) as
% 为什么要估计参数？ 使用least-square regression 工作区域 working distance 磁铁在S级表现对磁设备的吸引，而在N级表现排斥。 磁操纵的安全距离。
%The dipole moment modulus分别在全局和Omega距离内使用least-square regression以我们的实验结果进行了估计
\begin{equation}
	\label{eqn:optimalma}
	M_{\mc{A}}^\star=\arg\min_{M_{\mc{A}}}\left(\sum_{d\in\Omega}\big(B_{\mc{E}}(d)-B(d,M_{\mc{A}})\big)^2\right),
\end{equation}
where $B_{\mc{E}}$ represents experimental results and $B(d,M_{\mc{A}})$ is the scalar function of $d$ and $M_{\mc{A}}$, deduced from \eqref{eqn:magneticfield}. As a result, the red line with a reported value of $M_{\mc{A}}^\star=342.86$ $\mr{A\cdot\mr{m}^2}$ corresponds more closely to the FEM and experimental results within $\Omega$ than the globally optimal estimated. 

The spatial gradient magnetic field $\nabla\mb{b}\in\mathbb{R}^{3\times3}$ generated by the cylindrical magnet also contributes to actuate magnetic devices \cite{kimFerromagneticSoftContinuum2019}. The expression of $\nabla\mb{b}$ is derived from \eqref{eqn:magneticfield} as
% 由磁铁产生的空间梯度磁场对磁设备具有潜在的驱动能力。
\begin{equation}
	\label{eqn:gradientmagneticfield}
	\mr{\nabla}\mb{b}=\frac{3\mu_0M_{\mc{A}}}{4\pi\|\mb{p}\|^4}(\hb{p}\hb{m}_{\mc{A}}^\top+\hb{p}^\top\hb{m}_{\mc{A}}\mathbb{I}_3+\mr{Z}\hb{m}_{\mc{A}}\hb{p}^\top),
\end{equation}
where $\mr{Z}=\mathbb{I}_3-5\hb{p}\hb{p}^\top$. Appropriate quasi-static simplifications of Maxwell's equations ($\nabla\cdot\mb{b}=0$) and no current flow assumption ($\nabla\times\mb{b}=0$) in robotic applications limit the number of truly independent quantities in gradient magnetic field \cite{abbottMagneticMethodsRobotics2020}. The symmetry property is also reflected in \eqref{eqn:gradientmagneticfield} so that $\tfrac{\pt\mb{b}_{x}}{\pt y}=\tfrac{\pt\mb{b}_{y}}{\pt x}=0$. It can be observed in \figref{fig:MagnetAnalysis}.\textbf{C} that the red line presents good predictions for all three independent quantities of the gradient magnetic field along the principle axis in the distance $d$ within $\Omega$. However, the blue line with globally optimal estimated $M_{\mc{A}}$ appears to be more aggressive when compared with the FEM results. The reason might be that the point-dipole model takes no consideration of the magnet's geometric shape and an incomplete simplified resolution of the magnetic field is being used.

\subsection{Magnetic Soft Continuum Robot}

\begin{figure}[t]
	\centering
	\includegraphics[width=\columnwidth]{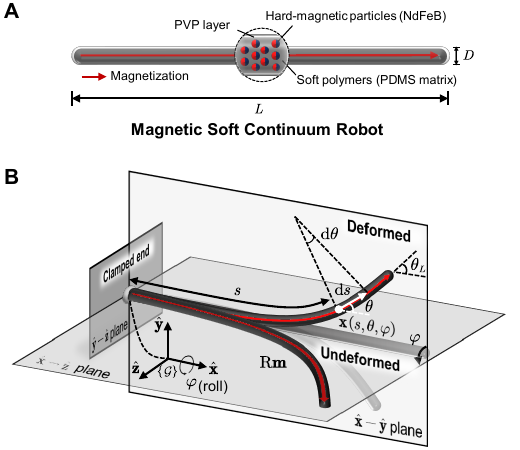}
	\caption{
Magnetic soft continuum robot (MSCR). (\textbf{A}) The material composition of the axially magnetized MSCR of length $L$ and diameter $D$. (\textbf{B}) A kinematic schematic of the MSCR in configuration space of $(s,\theta,\varphi)$.
 % Magnetic soft continuum robot (MSCR). (a) Material composition of the axially magnetized MSCR of length $L$ and diameter $D$. (b) A kinematic schematic of the MSCR in configuration space of $(s,\theta,\varphi)$, in which $s,\theta,\varphi$ represent the arc length, the tangential angle at the spatial material point $\mb{x}$, and roll angle, respectively. The rotation angle at the distal end is denoted in $\theta_L$. The magnetization vector $\mb{m}$ is premultiplied by a rotation matrix $\mr{R}$ in the deformed state.
}
	\label{fig:MSCR}
\end{figure}

The slender magnetic soft continuum robot, whose diameter $D$ is much less than its length $L$, is fabricated by injection molding with hard-magnetic soft composites where the fillers comprise soft polymers (PDMS matrix) and hard-magnetic particles ($6.5$-$\mu m$-sized NdFeB). As shown in \figref{fig:MSCR}.\textbf{A}, the lateral surface of the MSCR is encased by polyvinylpyrrolidone (PVP) thin layer, providing biocompatible, noncytotoxic, and hydrophilic properties in contact with vasculatures. Moreover, magnetic moments of the internal microparticles are rearranged axially under strong external magnetic fields, resulting in the isotropic and natural straight MSCR being magnetic manipulability in addition to super-elasticity.

The MSCR is modeled based on the hard-magnetic elastica theory \cite{wangHardmagneticElastica2020}, which 
parameterized it by a spatial curve $\theta=\theta(s)$ with an attached frame $\{\mathcal{G}\}$ at the clamped-proximal end. A schematic of the MSCR is demonstrated in \figref{fig:MSCR}.\textbf{B}. The arc length parameter $s$ and the tangential angle $\theta$ uniquely determine the spatial material point $\mb{x}(s,\theta)$ in planar deformed elastica. Furthermore, with the premise that torsion motion is negligible, the MSCR is capable of rolling along the $\hat{\mb{x}}_{\mc{G}}$-axis. Magnetic steering of the MSCR in planes with specific roll angles (noted as $\varphi$-plane) imposes constraints on the origin and orientation of the magnetic field generator, which will be discussed later. As a result, the configuration space of the MSCR is represented by $(s,\theta,\varphi)$, and the total Helmholtz free energy density consists of the following two parts:
\begin{equation}
\label{eqn:energypervolumn}
    \left\{\begin{aligned}
      &\varPsi^{\text{elastic}}(\theta^{\prime})=\frac{EI}{2A}\theta^{\prime^2}\\
      &\varPsi^{\text{magnetic}}(s, \theta; \varphi)=-\mr{R}\mb{m\cdot b},
    \end{aligned}\right.
\end{equation}
% &\varPsi^{\text{gravitational}}(s, \theta,\varphi)=\frac{\mathfrak{m}g}{AL}\hat{\mb{g}}\cdot \mb{x}=\rho g\hat{\mb{g}}\cdot\mb{x}
where $E, I, A$ represent Young's modulus, the second moment of inertia, and the cross-sectional area of the MSCR, respectively. The curvature $\kappa(s)=\tfrac{\mrd\theta(s)}{\mrd s}=\theta^\prime(s)$ forms a quadratic expression $\varPsi^{\text{elastic}}(\theta^\prime)$ of the elastic part. In respect to the magnetic part, the axial magnetization vector $\mb{m}\in\mathbb{R}^3$, referred to the undeformed state, is premultiplied by a rotation matrix $\mr{R}(\varphi,\theta)\in SO(3)$ in the $\varphi$-plane as $\mr{R}(\varphi,\theta)\mb{m} = \mr{Rot}(\hb{x}_{\mc{G}},\varphi)\mr{Rot}(\hb{z}_{\mc{G}},\theta)\mb{m}$,
where $\mr{Rot}(\cdot,\cdot)$ is known as Rodrigues' formula for rotations. The external magnetic field $\mb{b}(\mb{p},\hb{m}_{\mc{A}})$ is generated by the cylindrical magnet, in which $\mb{p}_{\mc{S}}$ is replaced by the material point $\mb{x}$ of interest. The total Helmholtz free energy density as a function of four variables $(s,\theta,\theta^\prime;\varphi):\varPsi(s,\theta,\theta^\prime;\varphi)=\varPsi^{\text{elastic}}(\theta^{\prime})+\varPsi^{\text{magnetic}}(s, \theta; \varphi)$. Then, the equilibrium state in arbitrary $\varphi$-planes can be described by the Euler-Lagrange equation: $\tfrac{\mrd}{\mrd s}\left(\tfrac{\pt\varPsi}{\pt\theta^\prime}\right)=\tfrac{\pt\varPsi}{\pt\theta}$, from which yields the kinematics, or the governing equation, of the MSCR as
\begin{equation}
	\label{eqn:governingequation}
	\frac{E I}{A} \frac{\mrd^2 \theta}{\mrd s^2}=-\frac{\pt}{\pt\theta}\left(\mr{R}\mb{m}\cdot \mb{b}\right).
\end{equation}
% +\rho g\hat{\mb{g}}\cdot\frac{\pt\mb{x}}{\pt\theta}
With the help of the chain rule, the right-hand side can be expanded as $\tfrac{\pt}{\pt\theta}\left(\mr{R}\mb{m\cdot b}\right)=\tfrac{\pt\mr{R}}{\pt\theta}\mb{m}\cdot\mb{b}+\left[\left(\nabla \mb{b}\right)^\top\mr{R}\mb{m}\right]\cdot\tfrac{\pt\mb{x}}{\pt\theta}$, where the material point $\mb{x}$ on the MSCR satisfies the following kinematic relations: 
\[
\mathbf{x}=\mathrm{Rot}(\hat{\mathbf{x}}_{\mathcal{G}},\varphi)\left[\int_0^s \cos \theta(\eta) \mrd \eta \hb{x}_{\mc{G}}+\int_0^s \sin \theta(\eta) \mrd \eta \hb{y}_{\mc{G}}\right]
\]
and
\[
\frac{\pt\mb{x}}{\pt\theta}=\mathrm{Rot}(\hat{\mathbf{x}}_{\mathcal{G}},\varphi)\!\left[-\!\!\int_0^s\!\!\sin \theta(\eta) \mrd \eta \hb{x}_{\mc{G}}\!+\!\int_0^s\!\!\cos \theta(\eta) \mrd \eta \hb{y}_{\mc{G}}\right]\!.
\]

To study the response of the MSCR's distal end under non-uniform magnetic fields, the governing equation \eqref{eqn:governingequation} is written in the canonical form with boundary conditions as
\begin{equation}
	\label{eqn:gvn_bvp}
	\frac{\mrd^2\theta}{\mrd s^2}=\sigma(s,\theta;\varphi).\quad \mbox{s.t.}\quad\theta(0)=0, \ \theta^\prime(L)=0,
\end{equation}
which is referred to as the two-point boundary-value problem (BVP). The crucial assumption that assists in establishing the existence of the solution $\theta(s)$ is presented as follows.
% 该问题的求解将要用到数值方法，例如shooting method或是collocation. 如下引理在确定BVP问题的解$\theta(s)$的存在性时是有用的。
% 我们首先给出帮助确定解$\theta(s)$的存在的重要引理和假设。
% The tangential angle $\theta$ along the arc length can be obtained by the integral equation:
% \begin{equation}
%     \theta(\xi)=\int_0^\xi\int_0^\zeta\sigma(s,\theta;\varphi)\mrd s\mrd\zeta.
% \end{equation}

%\begin{lemma}[Existence of solutions on BVP \cite{keller2018numerical}]
%    %\cite BVP
%    \label{lemma:extOfBVP}
%    Let the function $f(x,y,y^\prime)$ be continuous on $\mc{R}:a\leq x\leq b,y^2+{y^\prime}^2<\infty$, and satisfy there a uniform Lipschitz condition in $y$ and $y^\prime$. Then, the BVP constructed by $y^{\prime\prime}=f(x,y,y^\prime),\,\mbox{s.t.}\ a_0y(a)-a_1y^\prime(a)=\alpha,|a_0|+|a_1|\neq0;
%        b_0y(b)-b_1y^\prime(b)=\beta,\ |b_0|+|b_1|\neq0$, has solutions $y(x)$, which are sought on the interval $[a,b]\triangleq\{x|a\leq x\leq b\}$.
%\end{lemma}

% 
% The assumption are assumed to be hold
\begin{assumption}
    \label{ass:1}
    The position of the permanent magnet $\mb{p}_{\mc{A}}$ satisfies $\|\mb{p}_{\mc{A}}-\mb{x}(s)\|>0, \forall s\in[0,L]$.
\end{assumption}

\begin{theorem}
    \label{the:extTheta}
       The governing equation \eqref{eqn:gvn_bvp} for the MSCR has solutions $\theta(s)$ if the Assumption \ref{ass:1} is satisfied.
\end{theorem}

\begin{remark}
        Let $\|\mb{p}_{\mc{A}}-\mb{x}(s)\|\geq\|\mb{p}_{\mc{A}}\|-L>0$, this strong condition provides a feasibility scenario for Assumption \ref{ass:1} to hold. For practical implementations, the physical collision between the magnet and the MSCR requires consideration. One measure that can be taken is to ensure that the manipulation distance falls within the $\Omega$ interval designed in Section \ref{sec:II-A}.   % 在section II.A 中提出的操纵距离就显得至关重要
    \end{remark}
% Establishing the uniqueness of the solution $\theta(s)$ is challenging due to the highly nonlinear nature of $\sigma(s,\theta)$. 
\begin{remark}
    Some sufficient condition results on proofs of uniqueness have been made in the literature \cite{keller2018numerical}, indicating that the solution $\theta(s)$ is unique when $\tfrac{\pt\sigma(s,\theta)}{\pt\theta}>0$ in certain regions.
    % 1. 不提，2. 给出解的构造方式，3. 说明这是一个open problem
\end{remark}

\begin{figure}[t]
	\centering
	\includegraphics[width=0.9\columnwidth]{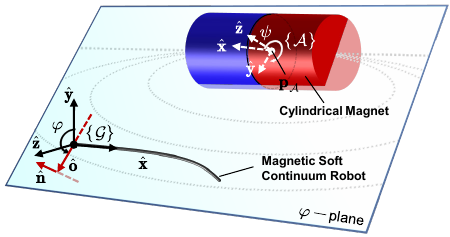}
	\caption{A feasible scheme for deflecting the MSCR in a specific plane. The geometry of the $\varphi$-plane is characterized by the normal $\hb{n}$ and the unit vector of intersection line $\hb{o}$ in frame $\{\mc{G}\}$. The cylindrical magnet with desired configuration $\mr{T}_{\mc{GA}}^d$ is located in the $\varphi$-plane and the rotation axis $\hb{z}_{\mc{A}}$ coincides with $\hb{n}$.}
	\label{fig:varphiplane}
\end{figure}

The proof of Theorem \ref{the:extTheta} is postponed to be completed in the Appendix.\ref{append:proof1} for readability. The solution to this problem will require the use of numerical methods, such as the shooting method or collocation. Let $\theta_L=\theta(L)$ be the rotation angle of the distal end, the integral equation of \eqref{eqn:gvn_bvp} reads
\begin{equation}
	\label{eqn:doubleIntegralforThetaL}
    \theta_L=\int_0^L\int_0^\zeta\sigma(s,\theta;\varphi)\mrd s\mrd\zeta=\vartheta(\mb{p}_{\mc{G}},\psi),
\end{equation}
where $\vartheta:\mathbb{R}^{3}\times\mathbb{R}\to\Theta$ represents the map from the control inputs to the workspace $\Theta$ of the MSCR. Recall that the configuration space (C-space) of the MSCR is represented by $(s,\theta,\varphi)$, of which the premise is that the robot always performs planar deflection. As illustrated in \figref{fig:varphiplane}, a feasible scheme is to constrain the origin of the magnetic field generator $\mb{p}_{\mc{A}}$ located in the $\varphi$-plane and to rotate only about the normal $\hb{n}$. Specifically, the unit vector of intersection line $\hb{o}$ and the normal $\hb{n}$ of the $\varphi$-plane are present in frame $\{\mc{G}\}$ as $\hb{o}_{\mc{G}} = [0,\cos\varphi,\sin\varphi]^\top$ and $\hb{n}_{\mc{G}} = [0,\sin\varphi,-\cos\varphi]^\top$. Then, the desired configuration of the cylindrical magnet is represented in frame $\{\mc{G}\}$ by the homogeneous matrix
$\mr{T}_{\mc{GA}}^d=(\mr{Rot}(\hat{\mb{n}}_{\mc{G}},\psi)\mr{R}_{\mc{GA}}^d(\varphi),\mb{p}_{\mc{GA}}^d)\in SE(3)$
, where $\mr{R}_{\mc{GA}}^d(\varphi)=[-\hb{x}_{\mc{G}},\hb{o}_{\mc{G}}(\varphi),\hb{n}_{\mc{G}}(\varphi)]$ constraints the orientation of the magnet, and $\mb{p}_{\mc{GA}}^d$ satisfies $\mb{p}_{\mc{GA}}^d\cdot\hb{n}=0$ to constraint the origin located in the $\varphi$-plane. The orientation matrix is premultiplied by $\mr{Rot}(\hb{n}_{\mc{G}},\psi)$, indicating the self-rotation of the magnet. In this manner, the principle axis of the rotating magnet stays in the plane with the deformed MSCR body, and, the distal end of the MSCR points to the S pole along symmetrical magnetic field streamlines. Moreover, if the origin $\mb{p}_{\mc{A}}$ is fixed, which means that the deformation of the MSCR is dominated by the rotation of the magnet, we are able to represent the C-space using only two variables $(\theta_L, \varphi)$.

\begin{remark}[On the C-space Representation]
\label{remark:On the C-space Representation}
Soft continuum robots are typically redundant and have theoretically infinite degrees of freedom (DOF). Because the control inputs are also redundant, independently moving or rotating the magnet can deflect the MSCR, making the map $\vartheta$ surjective to the inputs $\mb{p}_{\mc{G}}$ and $\psi$. Among all actuated configurations, those with the same $\theta_L$ but different deformed shapes constitute the null space $\mc{N}(\theta)$ for the tangential angle $\theta$. However, as pointed out in Theorem \ref{the:controllability}, the map $\psi\mapsto\vartheta(\cdot,\psi)$ is bijective when $\psi\in\Psi$, where the null space vanishes. Thus, the deformed MSCR actuated by rotating the magnet has a unique corresponding joint angle.

%Soft continuum robots are typically referred to as redundant and have theoretically infinite degrees of freedom (DOF). Because the control inputs are also redundant, independently moving or rotating the magnet is capable of deflecting the MSCR, the map $\vartheta$ is surjective to the inputs $\mb{p}_{\mc{G}}$ and $\psi$. Among all actuated configurations, ones with the same $\theta_L$ but different deformed shapes of the robot body constitute the null space $\mc{N}(\theta)$ for the tangential angle $\theta$. However, as we will point out in Theorem \ref{the:controllability}, the map $\psi\mapsto\vartheta(\cdot,\psi)$ is bijective in the condition that $\psi\in\Psi$, at which the null space vanished. Thus, the deformed MSCR actuated by rotating the magnet has a unique corresponding joint angle. 
\end{remark}
% characterization of 
% 有唯一对应的角度
% redundant discuss later, 可以看到连续的变化

% 冗余机器人，p ，psi构 establishes满射，相同的角度下构成零空间 然而 psi 双射，在下面叙述。因此对应确定状态，只用这两个描述。

% psi是单射，双射？
%+\frac{\pt\mathrm{Rot}}{\pt\theta}\mathbf{M}\cdot \mathbf{B}+\left((\operatorname{grad} \mathbf{B})^{\mathrm{T}} \mathrm{Rot}\mathbf{M}\right) \cdot \frac{\pt \mathbf{x}}{\pt \theta}-\rho g\hb{g}}\cdot\frac{\pt\mathbf{x}}{\pt\theta}

% 尖端始终指向S端的中心，因此导丝可以保持在过主轴的平面内运动

\section{Quasi-static Control Framework}
\label{sec:quasi-staticcontrol}

This section provides a quasi-static control framework for moving the MSCR's equilibrium states from one to another. The input-affine control system of the MSCR based on differential kinematics with analyzed Jacobian perturbation is primarily derived. Then, the linear extended state observer is employed to estimate the perturbation and provide robust control performance.

\subsection{Differential Kinematics}

To obtain the differential kinematic relation between the control inputs and the rotation angle of the distal end, the magnetic field \eqref{eqn:magneticfield} is re-represented as
\begin{equation}
	\label{eqn:separatedb}
	\mb{b}=\bar{\mr{B}}(\mb{p}_{\mc{A}})\hb{m}_{\mc{A}}(\psi).
\end{equation}
Note that the control inputs $\mb{p}_{\mc{A}}$ and $\psi$ are separated into a pair of product terms, the same rearrangement can be employed on the gradient magnetic field \eqref{eqn:gradientmagneticfield} by introducing an arbitrary vector $\mb{v}\in\mathbb{R}^3$, yielding the expression:
\begin{equation}
	\label{eqn:separatedgradientb}
	\begin{aligned}
		(\nabla\mb{b})^\top\mb{v}&=\frac{3\mu_0M_{\mc{A}}}{4\pi\|\mb{p}\|^4}(\hb{p}\hb{m}_{\mc{A}}^\top+\hb{p}^\top\hb{m}_{\mc{A}}\mathbb{I}_3+\mr{Z}\hb{m}_{\mc{A}}\hb{p}^\top)\mb{v}\\
		&=\frac{3\mu_0M_{\mc{A}}}{4\pi\|\mb{p}\|^4}(\hb{p}\mb{v}^\top+\mb{v}\hb{p}^\top+\hb{p}^\top\mb{v}\mr{Z})\hb{m}_{\mc{A}}\\
		&=\bar{\mr{B}}_{g}(\mb{p}_{\mc{A}},\mb{v})\hb{m}_{\mc{A}}(\psi).
	\end{aligned}
\end{equation}

Taking the derivative of $\theta(s,\theta;\psi)$ with respect to $\psi$ on \eqref{eqn:gvn_bvp} yields the Sturm-Liouville boundary-value problem (SL-BVP) that can be solved for the Jacobian $J(s)\in\mathbb{R}$ along the arc length:
    \begin{equation}
    \label{eqn:SLprob}
    \begin{aligned}
        &\frac{\mrd^2 J(s)}{\mrd s^2}=\frac{\pt\sigma(s,\theta)}{\pt \theta}J(s)+\frac{\pt\sigma(s,\theta;\psi)}{\pt\psi},\\
        &\mbox{s.t.}\quad J(0)=0,\ J^\prime(L)=0.
    \end{aligned}
    \end{equation}
    
To establish the existence and uniqueness of the solution to Jacobian $J(s)$, one can make the following assumption that ensures the Lipschitz constant $K$ derived in the proof of Theorem \ref{the:extTheta} is satisfied.
% 要确定Jacobian的解存在性和唯一性，可以通过让定理1的证明中推导的Lipschitz常数满足如下假设
\begin{assumption}
    \label{ass:2}
    The Lipschitz constant $K$ satisfies $K\in\mc{K}\triangleq[0,\frac{\pi^2}{4L^2}]\cup[(\frac{-\pi+4k\pi}{2L})^2,(\frac{\pi+4k\pi}{2L})^2],\ k\in\mathbb{N}$.
\end{assumption}

 \begin{theorem}
    \label{the:existenceOfSL-BVP}
        % SL-BVP的解始终存在，且当Assumption 2成立时，解是唯一的。
          SL-BVP \eqref{eqn:SLprob} has solutions $J(s)$ if Assumption \ref{ass:1} holds and it is unique when Assumption \ref{ass:2} holds.
\end{theorem}

 \begin{remark}
    \label{rmk:upperboundB}
    Limiting the magnitude of the magnetic field $B$ enables the Assumption \ref{ass:2} to hold. For $k=0$, the Lipschitz constant is subject to $0<K<\left(\frac{\pi}{2L}\right)^2$. The following numerical simulation results in Section \ref{subsec:jacobianAnalysis} permit us to disregard the norms of higher-order gradient tensors of $\mathbf{b}$, leading to an estimated upper bound as $B<\frac{EI\pi^2}{4MAL^2}$, which coincides with the critical field strength for MSCR buckling instability given in \cite{lu2023mechanics}.
\end{remark}
For the sake of readability, the proof of Theorem \ref{the:existenceOfSL-BVP} is located in Appendix \ref{append:proof2}. The infinite-dimensional distributed parameter system (DPS) formed by the Jacobian $J(s)$ provides a control model for any point on the MSCR body. However, in practical applications such as vascular interventions, the tip position and angle of MSCR are of greater concern. Controlling the isolated point ($s=L$) of the DPS system simplifies it into an input-affine nonlinear system. Without loss of generality, consider the scenario of $\varphi = 0$, the corresponding rotation matrix reads $\mr{R}_{\mc{GA}}(\varphi)=[-\hat{\mb{x}}_{\mc{G}},\hat{\mb{y}}_{\mc{G}},-\hat{\mb{z}}_{\mc{G}}$], and the unit dipole moment is given by $\hat{\mb{m}}_{\mc{G}}(\psi) = \mr{R}_{\mc{GA}}(\varphi)\hat{\mb{m}}_{\mc{A}}(\psi) = [-\cos\psi,\sin\psi,0]^\top$. Substituting the rearranged expressions \eqref{eqn:separatedb} and \eqref{eqn:separatedgradientb} into \eqref{eqn:doubleIntegralforThetaL} derives the inputs separated equation that $\vartheta(\mb{p}_{\mc{G}},\psi) = \bar{\vartheta}(\theta;\mb{p}_{\mc{G}})\hb{m}_{\mc{G}}(\psi)$, 
\begin{figure*}[b]
        \hrulefill
	\begin{equation}
		\label{eqn:separatedThetaL}
		\begin{aligned}
			\vartheta(\mb{p}_{\mc{G}},\psi)&=\left\{-\frac{A}{EI}\int_{0}^{L}\int_{0}^{\zeta}\left[\left(\frac{\pt\mr{R}}{\pt\theta}\mb{m}\right)^\top\bar{\mr{B}}(\mb{p}_{\mc{G}})+\left(\frac{\pt\mb{x}}{\pt\theta}\right)^\top\bar{\mr{B}}_{g}(\mb{p}_{\mc{G}},\mr{R}\mb{m})\right]\mrd s\mrd\zeta\right\}\hb{m}_{\mc{G}}(\psi)\\
			&=\bar{\vartheta}(\theta;\mb{p}_{\mc{G}})\hb{m}_{\mc{G}}(\psi).
		\end{aligned}
	\end{equation} 
        % \hrulefill
	% \vspace*{4pt}
\end{figure*}
where the details of $\vartheta(\mb{p}_{\mc{G}},\psi)$ are listed in \eqref{eqn:separatedThetaL}. If the origin of the magnet is fixed, $\bar{\vartheta}(\theta;\mb{p}_{\mc{G}})$ can be pre-computed. Consequently, $\hb{m}_{\mc{G}}$ is sorted out that facilitates deriving the partial derivative of $\theta_L$ with respect to $\psi$, which is 
\begin{equation}
	\label{eqn:parthetaLparpsi}
    \begin{aligned}
        \frac{\pt\theta_L}{\pt\psi}&=\frac{\pt\bar{\vartheta}(\theta;\mb
	p_{\mc{G}})}{\pt\theta}\frac{\pt\theta}{\pt\psi}\hb{m}_{\mc{G}}(\psi) + \bar{\vartheta}(\theta;\mb
	p_{\mc{G}})\frac{\pt\hb{m}_{\mc{G}}(\psi)}{\pt\psi}\\
    &=J(s)|_{s=L}\triangleq J_{\psi}(\mb{p}_{\mc{G}},\psi),
    \end{aligned}
\end{equation}
% Leibniz integral rule
where $J_{\psi}\in\mathbb{R}$ is the Jacobian of our interested, and $\tfrac{\pt\hb{m}_{\mc{G}}(\psi)}{\pt\psi}=[\sin\psi,\cos\psi,0]^\top$. 
% The latter term $\bar{J}_{\psi}$ is the predominant contributor of $J_\psi$, as will be indicated in the following. However, the former term $\Delta J_\psi$ is challenging to compute, yet it is bounded.
% A quantitative analysis of the bounds is presented in Section \ref{subsec:jacobianAnalysis}. 
Additionally, the partial derivative of $\theta_L$ with respect to $\mb{p}_{\mc{G}}$ is also considered for future reference. With the help of the expression: $\delta\mb{p}=\delta\mb{x}-\delta\mb{p}_{\mc{G}}$, it can be deduced that $\tfrac{\pt\theta_L}{\pt\mb{p}_{\mc{G}}}=-\tfrac{\pt\bar{\vartheta}(\theta;\mb{p}_{\mc{G}})}{\pt\mb{p}}\hb{m}_{\mc{G}}(\psi)=J_{\mb{p}}(\mb{p}_{\mc{G}},\psi)$, in which $\frac{\pt\bar{\vartheta}}{\pt\mb{p}}$ is a third-order tensor and $J_{\mb{p}}\in\mathbb{R}^{1\times 3}$. These results lead us to the differential kinematic equation: 
\begin{equation}
    \label{eqn:kine_eq}
    \dot{\theta}_L=J_{\psi}\dot{\psi}+J_{\mb{p}}\dot{\mb{p}}_{\mc{G}}.
\end{equation}

% \begin{remark}[On the Control Scheme of the MSCR]
%    Due to the redundancy of control inputs, our primary focus in this study lies in deflecting the MSCR by means of magnet rotation. However, future research will utilize magnet translation to achieve greater deflection of the MSCR.
% \end{remark}

\subsection{Analysis of the Jacobian}
\label{subsec:jacobianAnalysis}

\begin{figure*}[ht]
    \centering
    \includegraphics[width=\textwidth]{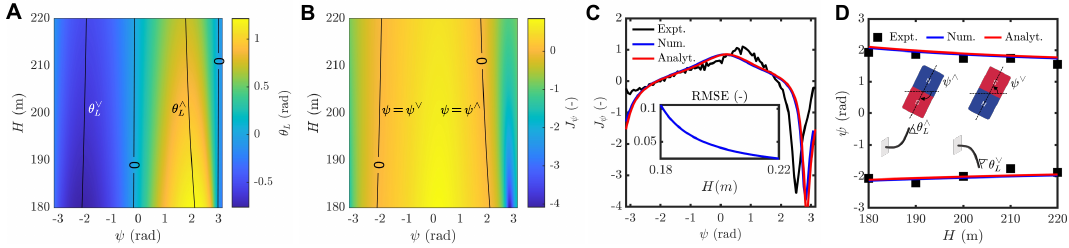}
    \caption{
    Heat map of (\textbf{A}) the rotation angle $\theta_L$ and (\textbf{B}) the Jacobian $J_\psi$ with the RME origin located at the distal end of the MSCR, varying in height $(H)$ from $0.18$ $\mr{m}$ to $0.22$ $\mr{m}$. (\textbf{C}) Comparison of the analytical Jacobian ($\bar{J}_\psi$) with the experimental result and numerical Jacobian when the RME positioned at $H=0.18 \mr{m}$ (see Movie \textcolor{blue}{S1}). (\textbf{D}) Comparison of the experimental, numerical, and analytical Jacobian singularities. 
    % Heat map of (a) the rotation angle $\theta_L$ and (b) the Jacobian $J_\psi$ with the RME origin located at the distal end of the MSCR, varying in height $(H)$ from $0.18$ $\mr{m}$ to $0.22$ $\mr{m}$. The workspace of $\theta_L$ is determined by the extremums $[\theta_L^\vee,\theta_L^\wedge]$ within the interval $\Psi=[\psi^\vee, \psi^\wedge]$ bounded by two extreme points. (c) Comparison of the analytical Jacobian ($\bar{J}_\psi$, red line) with the experimental result (black line) and numerical Jacobian (blue line) when the RME positioned at $H=0.18 \mr{m}$ (see Movie \textcolor{blue}{S4}). The fast convergence of the root-mean-square-error between the analytical and numerical results indicates $\bar{J}_{\psi}$ as a good estimator of $J_{\psi}$. (d) Comparison of the experimental (black square), numerical (blue line), and analytical (red line) Jacobian singularities. The inset demonstrates two Jacobian singularities corresponding to extreme points in the distal angle.
    }
    \label{fig:analysisJacobian}
\end{figure*}
% 插图展示了两处雅可比奇异，对应偏转角度的extremas.

% 证明了在满足约束条件下存在该区间

% 改成算法+分析
The analysis of the Jacobian $J_\psi(\mb{p}_{\mc{G}},\psi)$ is discussed in the special case of $\varphi=0$. Notably, this particular scenario holds for $\varphi\in[0, 2\pi]$. A preliminary qualitative analysis is conducted on the distal rotation angle $\theta_L$ of the MSCR, which is obtained from the product of $\bar{\vartheta}(\mb{p}_{\mc{G}})$ and $\hat{\mb{m}}_{\mc{G}}(\psi)$. The elements of $\bar{\vartheta}(\theta;\mb{p}_{\mc{G}})$ are represented as $[\bar{\vartheta}_1,\bar{\vartheta}_2,*]$. Subsequently, the expression of $\theta_L$ can be rewritten using the auxiliary angle formula as
\begin{equation}
\label{eqn:thetaLinsinewave}
    \begin{aligned}
        \theta_L =-\bar{\vartheta}_1\cos\psi+\bar{\vartheta}_2\sin\psi=\tilde{\vartheta}\sin(\psi-\psi_0),
    \end{aligned}
\end{equation}
where $\tilde{\vartheta}(\psi) = \sgn{\bar{\vartheta}_2}\sqrt{\bar{\vartheta}_1^2+\bar{\vartheta}_2^2}$, $\psi_0:=\psi_0(\psi)$ denotes the phase shift and $\psi_0=\arctan(\bar{\vartheta}_1/\bar{\vartheta}_2)$. 
The equivalence expression of $\theta_L$ implies that $\theta_L$ exhibits a periodic sinusoidal-like relation with $\psi$ and is bounded by $\tilde{\vartheta}$. These inferences are validated in the following quantitative computations. 
\begin{algorithm}[t]
    % \scriptsize
    \caption{Computation of deflection angle $\theta(s)$ and analytical Jacobian $J(s)$ of the MSCR}
    \label{algo:analytJacobian}
    \LinesNumbered
    \SetAlgoLined
    \SetAlgoNlRelativeSize{-1}
    \small
    \KwIn{Position $\mb{p}_{\mc{G}}$ and rotation angle $\psi$ of the magnet}
    \KwOut{Deflection angle $\theta(s)$ and analytical Jacobian $J(s)$}
    Compute $\theta(s)$ by solving the BVP (\ref{eqn:gvn_bvp}) using the shooting method \cite{ha2001nonlinear}\;
    Compute the initial value problems $LJ^{(1)}$ and $LJ^{(2)}$ with the higher order term $\nabla^2\mb{b}$ ignored using numerical integration methods\;
    \eIf{Assumption \ref{ass:2} is satisfied}{
    Linear combination factor $v=-J^{(1)^{\prime}}(L)/J^{(2)^\prime}(L)$\;
    Analytical Jacobian $J(s)=J^{(1)}(s)+vJ^{(2)}(s)$\;
    }{
    Analytical Jacobian $J(s)=J^{(1)}(s)$\;
    }
\end{algorithm}

The deflection angle $\theta(s)$ and the analytical Jacobian $J(s)$ can be obtained by implementing the Algorithm \ref{algo:analytJacobian}. In the shooting method, the second derivative $\tfrac{\partial^2\sigma(s,\theta)}{\partial\theta^2}$ used for iteration is estimated through Newton interpolation, leading to efficient convergence with an average number of shooting iterations less than 3 times. Combining \eqref{eqn:magneticfield} and the conclusion in Remark \ref{rmk:upperboundB}, Assumption \ref{ass:2} holds when the position between the magnet and the robot satisfies the resulting inequality:
\begin{equation}
    \label{eqn:max_p}
    \|\mb{p}\|>\frac{1}{\pi}\left(\frac{\mu_0M_\mc{A}MAL^2}{EI}\right)^{\frac{1}{3}}.
\end{equation}
With the material property of MSCR \#1 lied in Tab. \ref{tab:parameterOfMSCR}, the upon inequality shows $\|\mb{p}\|>0.1425\ (\mr{m})$. By locating the RME at the top of the MSCR's distal end with height varying from $0.18$ $\mr{m}$ to $0.22$ $\mr{m}$, a heat map illustrating the values of $\theta_L$ is demonstrated in \figref{fig:analysisJacobian}.\textbf{A}.
% Then, by Theorem \ref{the:existenceOfSL-BVP}, there exists a unique solution to problem \eqref{eqn:SLprob} when the height $H$ 
% It provides an alternative equivalent expression of $J_{\psi}$ by taking the partial derivative of $\psi$ on both sides of the upon equation:
% % 这与公式(19)有什么不同？公式(19)是可以用于计算的式子，这里仅是定性表达式，用来求解
% \begin{equation}
% 	\label{eqn:quant_Jacobian}
% 	\frac{\pt\theta_L}{\pt\psi} = \tilde{\vartheta}\cos(\psi-\psi_0)\left(1-\frac{\pt{\psi_0}}{\pt\psi}\right)+\frac{\pt\tilde{\vartheta}}{\pt\psi}\sin(\psi-\psi_0).
% \end{equation}
% Comparing to the previously derived Jacobian \eqref{eqn:parthetaLparpsi}, the qualitative expression of $\bar{J}_\psi$ is extracted as $\bar{J}_{\psi} = \tilde{\vartheta}\cos(\psi-\psi_0)$, and the pertubative term $\Delta J_{\psi}$ is bouned by
% \begin{equation}
%     \label{eqn:deltaJacobian}
%     \begin{aligned}
%         |\Delta J_\psi|&=\left|-\frac{\pt{\psi_0}}{\pt\psi}\tilde{\vartheta}\cos(\psi-\psi_0)+\frac{\pt\tilde{\vartheta}}{\pt\psi}\sin(\psi-\psi_0)\right|\\
%     &\leq \sqrt{\left(\frac{\pt\psi_0}{\pt\psi}\tilde{\vartheta}\right)^2 + \left(\frac{\pt\tilde{\vartheta}}{\pt\psi}\right)^2}.
%     \end{aligned}
% \end{equation}
The numerical results indicate that as $\psi$ varies within the range of $[-\pi, \pi]$, the sinusoidal-like trend of $\theta_L$ is well predicted by \eqref{eqn:thetaLinsinewave}. Similarly, the successfully predicted phase shift contour ($\theta_L= 0$) explains the deflection of MSCR induced by the gradient magnetic field when $\psi=0$. The workspace of the MSCR is determined by the extremums, denoted as $\theta_L^\vee$ and $\theta_L^\wedge$, and it becomes compressed as the RME moves away. The proof of the boundedness and bijection property of $\theta_L$ is scheduled in Theorem \ref{the:controllability}. The extreme points $\psi^\vee$ and $\psi^\wedge$ are solutions such that $\frac{\pt\theta_L}{\pt\psi}(\psi) = 0$. Despite the displacement of the extremums of MSCR at different heights, its periodicity of $2\pi$ remains unchanged. As depicted in \figref{fig:analysisJacobian}.\textbf{B}, the zero-crossing contour lines of the Jacobian $J_\psi$ correspond to the extremal trajectories of $\theta_L$, and highlighting the singular conditions. 

Recall that $J(s)$ is computed with higher order term $\nabla^2\mb{b}$ ignored, a comparison between the experimental, numerical, and analytical Jacobian at $H=0.18\ \mr(m)$ is plotted in \figref{fig:analysisJacobian}.\textbf{C}. The results indicate that the analytical Jacobian (denoted as $\bar{J}_{\psi}$) fits well with the numerical Jacobian (noted as $\Delta J_{\psi}$), with the root-mean-square-error (RMSE) caused by $\nabla^2\mb{b}$ terms decreasing as $H$ increases. Both of them exhibit a good agreement with experimental results and explain the rapid reduction in the deflection angle of MSCR when the magnet crosses the greater singular value $\psi^\wedge$. As shown in Movie \textcolor{blue}{S1}, the magnet briefly loses control over MSCR when crossing $\psi^\wedge$ and regains control after further rotating. This transient process can be described by the dynamic model proposed in our previous work \cite{wu2023kirchhoff}. \figref{fig:analysisJacobian}.\textbf{D} illustrates the singularities predicted by our model for the magnet at different heights. Our model yields RMSE of $0.1225\ (\mr{rad})$ for $\psi^{\wedge}$ and $0.1595\ (\mr{rad})$ for $\psi^{\vee}$. These accurate estimates are beneficial for precisely determining the control direction and allow the addition of soft constraints to limit the rotation range for RME, thereby preventing loss of control over the MSCR.
% 结果表明，解析Jacobian (noted in $J\bar(\psi)$)对数值Jacobian的拟合效果很好，由包含$\nabla^2\mb{b}$的项（noted in $\Delta J\bar(\psi)$造成的RMSE随着$H$的增大而快速下降。二者都展现了与实验结果的良好的一致性，并且解释了当磁铁的转角越过较大奇异位置后，MSCR的偏转角度的快速回落现象。如Movie S5中展示的那样，越过奇异位置的磁铁短暂的失去了对MSCR的控制权，并在进一步的调整旋转角度$\psi$后恢复。该瞬态过程可以由我们之前的工作中提出的动力学模型\cite{wu2023kirchhoff}所描述。\figref{fig:analysisJacobian}.(d)展示了我们的模型预报的磁铁位于不同高度下的奇异值。这一良好的估计结果有利于精准确定控制方向，并且允许为RME添加软约束来限制旋转范围，从而避免丢失对MSCR的控制。

\subsection{Design of the Controller}

% \begin{figure}[t]
%     \centering
%     \includegraphics[width=0.9\columnwidth]{figures/Fig5.pdf}
%     \caption{Closed-loop deflecting control diagram of the MSCR.}
%     \label{fig:controldiagram}
% \end{figure}
 
A closed-loop control scheme is designed for the MSCR to perform quasi-static state transfer and can overcome the model parameter perturbation $\Delta J_{\psi}$. Define the system's state as $x_1=\theta_L$, the input as $u=\dot{\psi}$, and $x_1(t)$ directly as the output $y(t)$. Then, the differential kinematic model is rewritten by
\begin{equation}
    \label{eqn:system_model}
    \left\{
    \begin{aligned}
        \dot{x}_1(t) &= x_2(t) + \bar{J}_\psi u(t)\\
        y(t) &= x_1(t),
    \end{aligned}\right.
\end{equation}
where $x_2(t)\triangleq f(x_1(t), w(t))+\Delta J_{\psi}u(t)$ denotes the extended state covering model uncertainties, perturbances, and unknown disturbances. The controllability of this affine nonlinear system is described by the following theorem.
\begin{theorem}
\label{the:controllability}
    If Assumptions \ref{ass:1} and \ref{ass:2} are hold, then the distal rotation angle $\theta_L=\vartheta(\psi)$ exists a value domain $\Theta\triangleq[\theta_L^\vee,\theta_L^\wedge]$ when $\psi$ lies in an arbitrarily closed interval $\Psi\subset\mathbb{R}$, and the state $x_1(t)=\theta_L$ of the system \eqref{eqn:system_model} is controllable for $\theta_L\in\Theta$.
     % system 是局部可控的
\end{theorem}

\begin{remark}
    Notably, the unit vector $\mathbf{m}_{\mathcal{G}}(\psi) = [\cos(\psi), \sin(\psi), 0]^\top$ exhibits a period length of $2\pi$. Thus, when the length of $\Psi$ exceeds $2\pi$, $\theta_L$ will encompass a widthest domain of $\Theta$. Another important result from the above theorem is that $\theta_L\in\Theta$ forms a monotonic bijective mapping with $\psi$, in response to Remark \ref{remark:On the C-space Representation}.
\end{remark}
The proof of Theorem \ref{the:controllability} is provided in Appendix \ref{append:proof3}. As thoroughly elaborated in \cite{hanPIDActiveDisturbance2009}, the extended state observer (ESO) is well-constructed to estimate the total disturbance $x_2(t)$. To apply conveniently in practice, a linear ESO (LESO) is employed in the input-affine system \eqref{eqn:system_model} as
% linear ESO
\begin{equation}
    \label{eqn:LinearESO}
    \left\{
    \begin{aligned}
        \dot{\bar{x}}_1(t) &= \bar{x}_2(t) + \frac{\beta_1}{\epsilon}(y(t) - \bar{x}_1(t)) + \bar{J}_\psi u(t)\\
        \dot{\bar{x}}_2(t) &= \frac{\beta_2}{\epsilon^2}(y(t) - \bar{x}_1(t)),
    \end{aligned}\right.
\end{equation}
where $\beta_1, \beta_2$ are two pertinent observer gains, and $\epsilon\geq0$ is the constant gain. A proportional-derivative (PD) controller is designed as $u_0(t) = \dot{\bar{y}}_r(t)+k(\bar{y}_r(t)-\bar{x}_1(t))$, where $\bar{y}_r(t)$ and $\dot{\bar{y}}_r(t)$ are the tracked and the derivative of the reference signal $y(t)$ using the tracking differentiator (TD), respectively. A useful linear structure of TD is given by 
$\ddot{\bar{y}}_r(t) = -k_1R^2(\bar{y}_r(t) - y_r(t))-k_2R\dot{\bar{y}}_r(t)$, where $k_1>0, k_2>0$ are constants, and $R>0$ is the tuning parameter. Note that the stability analysis and the convergence proof of LESO and TD are vividly outlined in \cite{guoActiveDisturbanceRejection2016}. With the extended state $\bar{x}_2(t)$ uniformly convergence to $x_2(t)$, the comprehensive controller that cancels the total disturbance is designed as
\begin{equation}
    \label{eqn:LADRCcontroller}
	u(t)=\bar{J}_{\psi}^{-1}(u_0(t)-\bar{x}_2(t)).
\end{equation}
It can be readily verified that the upon controller exponentially converges to the reference signal over time. The feedback loop is accomplished by the visual servoing platform, clarified in the following Section \ref{subsec:visualServo}.
% visual servo create algorithm feedback loop

\section{Simulations and Experiments}
\label{sec:experiment}
\begin{figure}[t]
    \centering
    \includegraphics[width=\columnwidth]{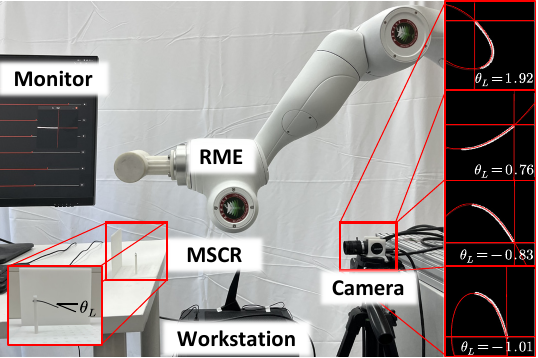}
    \caption{A photo shot of the experiments with instrumentation. The distal rotation angle of the MSCR is measured by conic section fitting.}
    \label{fig:systemOverview}
\end{figure}
% 视觉伺服获得MSCR的尖端偏转角度
This section validates the feasibility of the previously proposed quasi-static control framework. A fast and precise algorithm is developed and deployed on the visual servoing platform to capture the distal rotation angle. The change in control direction introduced by $J_\psi$ is demonstrated to be necessary in the simulation. The effectiveness of the closed-loop deflection control using PD and quasi-static controller (QSC) is compared. Experimental procedures are provided in the \textit{\textcolor{blue}{Supplementary Videos}}.
% 这一小节验证了之前提出的准静态控制框架的可行性。基于视觉伺服，设计了快速鲁棒的算法用于捕获远端旋转角。 对比了开闭环偏转控制的控制效果。实验过程在支撑视频中给出。

% 引入$J_\psi$所带来的控制方向改变在仿真中被证明是必要的。

\subsection{System Overview}

A photo shot of the experiments with instrumentation is displayed in \figref{fig:systemOverview}. The high-performance workstation (Intel$^\circledR$ Core\textsuperscript{TM} i5-12490F @3.00GHz processor and 32GB RAM) was utilized for computing control commands and device communication. The system devices are comprised of several components as follows:
\begin{inparaenum}[(1)]
\item Magnetically actuated MSCRs were magnetized along the robot's axial direction using a magnetizer (MA-3030, Jiuju Inc., Shenzhen, China).
\item The customized 3D-printed robot fixture platform was employed to clamp the proximal end of the robot and allow flexible configuration of the free end length.
\item Cameras (HKMV-CH050-10VC, Hangzhou Hikrobot Co., Ltd., China) were utilized for visual servoing to achieve closed-loop control of the robot's deflection.
\item  The RME was mounted at the end-effector of a robot arm (Diana 7w, AGILE ROBOTS AG, Germany) to ensure reaching desired configurations.
\end{inparaenum}
% Zhenxing Yang 为什么添加了一段实验设备的定性描述？

% \begin{table}[ht]
% \begin{center}
% \caption{Parameters of the MSCRs.}
% \label{tab:parameterOfMSCR}
% \begin{tabular}{| c | c | c |}
% \hline
% Parameter (Description) & Value & Unit\\
% \hline
% $L$ (Length) & $24$ & $\mr{mm}$ \\
% \hline
% $r$ (Radius) & $0.54$ & $\mr{mm}$\\ 
% \hline
% $E$ (Young's modulus) & $3.0$ & $\mr{MPa}$\\
% \hline 
% $M$ (Magnitude of magnetization) & $8.0$ & $\mr{kA/m}$\\
% \hline
% \end{tabular}
% \end{center}
% \end{table}

%\begin{table}[t]
%	\begin{center}
%		\caption{Parameters of the MSCRs.}
%		\label{tab:parameterOfMSCR}
%		\begin{tabular}{clll}
%			\hline
%			\rowcolor{gray} 
%			\multicolumn{1}{l}{\cellcolor{gray}} & \textbf{Parameter (Description)} & \textbf{Value} & \textbf{Unit} \\ \hline
%			& $L$ (length) & 24 & $\mathrm{mm}$ \\
%			& $r$ (Radius) & 0.54 & $\mathrm{mm}$ \\
%			& $E$ (Young's modulus) & 3.0 & $\mathrm{MPa}$ \\
%			\multirow{-4}{*}{MSCR \#1} & $M$ (Magnitude of magnetization) & 8.0 & $\mathrm{kA/m}$ \\ \hline
%			& $L$ (length) & 30 & $\mathrm{mm}$ \\
%			& $r$ (Radius) & 0.65 & $\mathrm{mm}$ \\
%			& $E$ (Young's modulus) & 2.8 & $\mathrm{MPa}$ \\
%			\multirow{-4}{*}{MSCR \#2} & $M$ (Magnitude of magnetization) & 9.3 & $\mathrm{kA/m}$ \\ \hline
%		\end{tabular}
%	\end{center}
%\end{table}

\begin{table}[t]
\begin{center}
\caption{Parameters of the MSCRs.}
\label{tab:parameterOfMSCR}
\begin{tabular}{clll}
\hline
\rowcolor[HTML]{EFEFEF} 
\multicolumn{1}{l}{\cellcolor[HTML]{EFEFEF}} & Parameter (Description) & Value & Unit \\ \hline
 & $L$ (length) & 24 & $\mathrm{mm}$ \\
 & $r$ (Radius) & 0.54 & $\mathrm{mm}$ \\
 & $E$ (Young's modulus) & 3.0 & $\mathrm{MPa}$ \\
\multirow{-4}{*}{MSCR \#1} & $M$ (Magnitude of magnetization) & 8.0 & $\mathrm{kA/m}$ \\ \hline
 & $L$ (length) & 30 & $\mathrm{mm}$ \\
 & $r$ (Radius) & 0.65 & $\mathrm{mm}$ \\
 & $E$ (Young's modulus) & 2.8 & $\mathrm{MPa}$ \\
\multirow{-4}{*}{MSCR \#2} & $M$ (Magnitude of magnetization) & 9.3 & $\mathrm{kA/m}$ \\ \hline
\end{tabular}
\end{center}
\end{table}  

\subsection{Visual Servoing Platform}
\label{subsec:visualServo}

The measurement of the MSCR's distal angle $\theta_L$ is challenging due to the lack of body-load sensors. Currently, visual servoing and emerging magnetic/ultrasonic localization techniques are the mainstream measurement methods \cite{yangUltrasoundGuidedCatheterizationUsing2022,zhang5DLargeWorkspaceMagnetic2023,sikorskiVisionBased3DControl2019}. Image-based visual servoing fits conic sections to the MSCR to capture $\theta_L$. This post-processing procedure is outlined in Algorithm \ref{algo:measureOfdistalangle}. The algorithm requires real-time images of the MSCR, typically obtained through deep learning or threshold segmentation, containing continuous pixels from the proximal to the distal end. The core idea is to obtain the analytical expression of the fitting function at different deformation shapes using LSF, then iteratively compute the pixels or their neighborhood from a starting point to determine if the endpoint is reached. Finally, the derivative at the end-point, representing the tangent slope corresponding to the deflection angle $\theta_L$, is calculated.

%The measurement of the MSCR's distal angle $\theta_L$ is challenging due to the unavailability of body-load sensors. Currently, visual servoing and emerging magnetic/ultrasonic localization techniques are among the mainstream measurement methods \cite{yangUltrasoundGuidedCatheterizationUsing2022,zhang5DLargeWorkspaceMagnetic2023,sikorskiVisionBased3DControl2019}. The image-based visual servoing allows us fitting conic sections to the MSCR for capturing $\theta_L$. This image post-processing procedure is organized into Algorithm \ref{algo:measureOfdistalangle}. The algorithm requires real-time images captured by the camera that contain continuous pixels of the MSCR from the proximal to the distal end, typically obtained through deep learning or threshold segmentation. The core idea of the algorithm is first to obtain the analytical expression of the fitting function/equation of the MSCR at different deformation shapes based on LSF, and then iteratively compute the pixel or its neighborhood from a starting point to determine whether the endpoint is reached. Finally, the derivative at the endpoint is calculated based on the analytical expression, representing the tangent slope corresponding to the deflection angle $\theta_L$. 

We utilize the linearity measure $\ell_e$ to quantify deformation and select the appropriate fitting function. The starting point is based on the image centroid $P$, ensuring it remains valid within the robot body. The algorithm results are shown in \figref{fig:systemOverview}. For small deflections, a quadratic function is chosen to improve computational speed, with coordinates as the iterative variable. For large deflections, an elliptical equation is used, with the eccentric angle as the iterative variable. By transforming coordinates to the ellipse center, iterative computation is simplified using the standard elliptic parametric equation. The algorithm achieves a frame rate exceeding $75$ $\mr{Hz}$, capturing the dimensionless distal rotation angle of the MSCR in any configuration without markers on the free end.

%We utilize the linearity measure, denoted as $\ell_e$, to quantify the degree of deformation and choose the appropriate fitting function/equation. The selection of the starting point is determined based on the centroid of the image $P$, ensuring that the starting point remains within the robot body and is valid. The results of the algorithm are presented in \figref{fig:systemOverview}. For small deflections, a quadratic function is chosen to improve computational speed, and the coordinates are regarded as the iterative variable. For large deflections, an elliptical equation is employed, and the eccentric angle is selected as the iterative variable. By performing a coordinate transformation to the center of the ellipse, the iterative computation is simplified using the standard elliptic parametric equation. The algorithm achieves a frame rate exceeding $75$ $\mr{Hz}$, enabling the capture of the dimensionless distal rotation angle of the MSCR in any configuration without any markers on the free end.

\begin{algorithm}[h]
        \scriptsize
	\caption{Conic section fitting for distal angle}
	\label{algo:measureOfdistalangle}
	\LinesNumbered
    \SetAlgoLined
    \SetAlgoNlRelativeSize{-1}
    \small
	\KwIn{Real-time binarized image $\mr{P}\in\mathbb{R}^{u\times v}$} % MSCR的实时二值化图像P 
	\KwOut{Conic section fitted distal angle $\theta_L$}
    Compute the centroid pixel coordinates $(g_x, g_y)$ of $\mr{P}$\;
	Compute the linearity $\ell_e$ of the MSCR\;
    \eIf{$\ell_e\leq $ threshold }{
        \tcp{small deflection}
        LSF using the quadratic function: $\bar{v}(\bar{u}) = a^2\bar{u}+b\bar{u}+c$\;
        Set the iteration initial point at $t_x=g_x, t_y=\bar{v}(g_x)$\;
        \Do{$\mr{P}([t_x],[t_y])=0$}{
        $t_y\gets t_y\pm2at_x+a^2\pm b$, $t_x\gets t_x\pm1$\;
        Compute the pixel at $([t_x],[t_y])$ (or the sum of its $3\times 3$ neighborhood for robustness)\;
        }
    }{
        \tcp{large deflection}
        LSF using the elliptical function:
        $(\bar{u}, \bar{v})\mr{Q}(\bar{u}, \bar{v})^\top$\;
        Coordinates transformation from camera to ellipse $\{O_{\mc{E}}\}$\;
        Set the iteration initial angle $\alpha=\arctan(g_y^\mc{E}/g_x^\mc{E})$\;
        % Compute the slope angle of the origin-barycenter line as the start angle $\alpha$, and the start point using the parametric equation  \;
        \Do{$\mr{P}([t_x],[t_y])=0$}{
        $\alpha\gets \alpha\pm0.1, t_x^{\mc{E}} = a\cos\alpha, t_y^{\mc{E}} = b\sin\alpha$\;
        \tcp{standard elliptic parametric equation}
        Compute the pixel at $([t_x],[t_y])$ (or the sum of its $3\times 3$ neighborhood for robustness)\;
        }
        % $a\bar{u}^2+b\bar{u}\bar{v}+c\bar{v}^2+d\bar{u}+e\bar{v}+f=0$\;
    }
    Compute the slope angle at $([t_x],[t_y])$ and select the one with the larger absolute value for $\theta_L$\; 
\end{algorithm}

% 1. 计算图像重心（G_x,G_y）
% 2. 计算与线性函数拟合的均方根误差 l_e
% 3.1 如果l_e小于阈值（小挠度形变）
% 3.1.1 与二次函数y =  ax^2+bx+c 拟合
% 3.1.2 选取起始点为G_x, y(G_x)
% 3.1.3 do
% 3.1.4 G_x<-G_x+1，更新y
% 3.1.5 计算该点3x3邻域像素
% 3.1.6 while 无像素 // 尖端点
% 3.2 否则l_e大于阈值 （大挠度形变）
% 3.2.1 与椭圆ax^2+bxy+cy^2+dx+ey+f=0拟合
% 3.2.2 坐标变换至椭圆圆心，建立参数方程
% 3.2.3 选取起始角度为圆心与重心连线的离心角e_\theta
% 3.2.4 do
% 3.2.5 e_\theta<-e_\theta+1/-1
% 3.2.6 计算新的交点3x3邻域像素
% 3.2.7 while 无像素 // 尖端点/起始点
% 4 计算尖端点处拟合函数的导数，取绝对值较大的atan作为尖端角度

\subsection{Experimental Results}

\begin{figure*}[ht]
    \centering
    \includegraphics[width=0.9\textwidth]{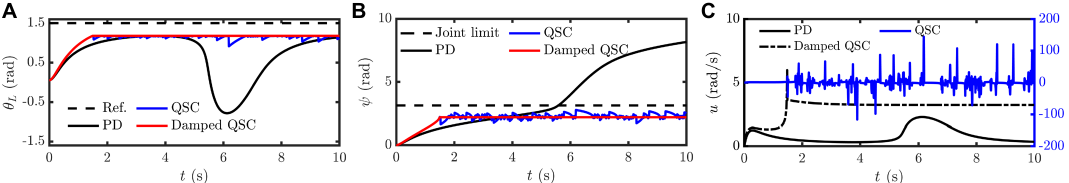}
    \caption{(\textbf{A}) Comparison of the responses of an unreachable step signal by the PD controller, quasi-static controller (QSC), and damped QSC. (\textbf{B}) Comparison of the rotation angle $(\psi)$ of the actuator. Note that the control law generated by the PD controller results in the angle exceeding the joint limit. (\textbf{C}) Comparison of the control law. The chattering in the QSC controller is caused by discontinuities in the Jacobian near singularity.}
    \label{fig:controldirection}
\end{figure*}
% 不可达阶跃信号的响应对比，使用
% 注意PD控制器产生的控制律导致角度超过了关节极限

\begin{figure}[ht]
    \centering
    \includegraphics[width=\columnwidth]{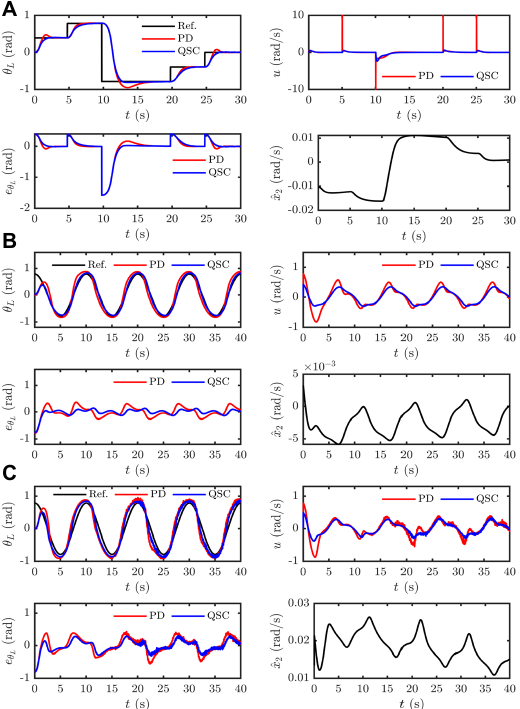}
    \caption{Comparison of the (\textbf{A}) step-response, (\textbf{B}) cosine-response, and (\textbf{C}) cosine-response with disturbance of MSCR using the QSC and PD controller. The transient processes, control inputs, and estimated total disturbances are plotted in each panel.}
    \label{fig:expresults}
\end{figure}

% 实验设置：模型参数总结在Tab. 1中，名义Jacobian的计算。机器人固定端长度
The measured physical parameters of the MSCR \#1 are summarized in Tab. \ref{tab:parameterOfMSCR}. The nominal Jacobian $\bar{J}_\psi$ in \eqref{eqn:LADRCcontroller} is obtained by carrying out Algorithm \ref{algo:analytJacobian}. Although our designed control strategy constrains the configuration of the RME, it also enables the three-dimensional deflection of the MSCR to be equivalently represented as a two-dimensional deflection in the $\varphi$-plane. Due to the dual operations of attraction and repulsion exerted by the permanent magnet on the MSCR, in situations where the RME is unable to reach certain configurations (e.g., physical collision constraints or joint singularities), tasks can be accomplished by the dual operation from the opposite side. For the sake of simplicity in measuring the distal angle, we select the $\hb{x}_{\mathcal{G}}$-$\hb{y}_{\mathcal{G}}$ plane to validate the performance of the quasi-static controller. During the experimental process, the origin of the RME remains fixed above the distal end of the MSCR at the height of $18.0$ $\mr{cm}$, and the initial rotation angle is set to zero. However, because of the existing gradient magnetic field, the MSCR undergoes slight deflections as shown in \figref{fig:analysisJacobian}.\textbf{A}, which gradually diminishes as the RME moves further away. The initial value of $\theta_L$ is then set to zero, and as demonstrated in Movies \textcolor{blue}{S2-S4}, the rotation angle of the RME stabilizes close to $\varphi_0$. The parameters of the LESO are designed as $\beta_1 = 1$, $\beta_2 = 0.01$, and $\epsilon = 0.01$, while the parameters of TD are set as $R=10$, $k_1=0.1$, and $k_2=1$. As for the PD controller, $k=1.02$.

We first validate the necessity of introducing the change in control direction brought about by $J_\psi$ through simulations. Previous analysis has indicated that the system is bounded under any initial conditions. In practical applications, it is difficult to avoid setting reference signals that extend beyond the system's state range $\Theta$. Therefore, when the actuator is near the extreme points $\psi^{\wedge}$ or $\psi^{\vee}$, the sign of $J_{\psi}$ may change and potentially cause singular issues in the system. A straightforward solution inspired from \cite{buss2005selectively} is to introduce a damping form for the structural change of $J_\psi$ as follows:
\begin{equation}
    J_{\psi} = \begin{cases}
        J_{\psi}, &|J_{\psi}|\geq\lambda\\
        \mbox{sgn}(J_{\psi})\lambda, &|J_{\psi}|<\lambda,
    \end{cases}
\end{equation}
where $\lambda$ is a small positive constant. We refer to the QSC with this structure introduced as the damped QSC. This approach directly prevents the system from velocity singularities when $J_\psi$ becomes singular.
% 我们设计的控制策略虽然约束了RME的构型，但是能够将MSCR的三维偏转等效为在$\phi$-平面内的二维偏转。由于永磁铁对MSCR存在吸引和排斥两种操作模式，对于RME难以到达的构型（例如由于物理碰撞限制，或关节奇异等），可以采用在对侧吸引/排斥的方式完成任务。需要注意的是，我们假设扭矩始终是忽略不计的。因此，为了简便尖端角度的测量，我们选择在$\hb{x}_{\mc{G}}-\hb{y}_{\mc{G}}$平面验证准静态控制器的性能。实验过程中，RME的origin始终固定在MSCR远端的上方，测量的高度为$18.6$ $\mr{cm}$，初始化旋转角度被设置为0。然而，存在的梯度磁场迫使MSCR发生轻微的偏转，如图\figref{} .(a)所示，这一现象随着RME远离逐渐消失。$\theta_L$的初始值被紧接着设置为0，如Movie S1-S3中所展示的那样，RME的初始角度最终稳定在$\phi_0$。LESO的参数设计为$\beta_1 = 1,\beta_2 = 0.01,\epsilon = 0.01$, TD参数设计为$R=10,k_1=0.1,k_2=1$, PD控制器中$k=1.0$。

In \figref{fig:controldirection}.\textbf{A}, the system's response to an unreachable step signal using PD, QSC, and damped QSC controllers is shown. The PD controller, lacking the correct control direction, produces residual errors that persist after extremum points, forcing continuous actuator operation. As seen in \figref{fig:controldirection}.\textbf{B}, this incorrect control direction causes the actuator to reach the joint limit around 5.8 seconds, posing safety risks. In contrast, QSC, with the correct control direction, stabilizes the system near output limits, keeping the actuator within safe joint limits. However, Jacobian singularities cause high-frequency chattering, leading to the spiky output seen in \figref{fig:controldirection}.\textbf{C}. Damped QSC mitigates this chattering, stabilizing the system near extremums with a smooth control law and safe actuator operation. In summary, the control direction introduced by $J_{\psi}$ protects system components and enhances performance.

Within the workspace of the MSCR, a comparison is conducted between the step responses of the MSCR using the PD controller and QSC approach (see Movie \textcolor{blue}{S2}). As shown in \figref{fig:expresults}.\textbf{A}, under the same parameter conditions, the QSC approach significantly eliminates overshoot while maintaining a fast response. This improvement can be attributed to the reliable nominal controller gains provided by $\bar{J}_{\psi}$, regulating the controller parameters, and the incorporation of the ESO for estimating unmodeled dynamics $\bar{x}_2$. The performance error plot illustrates the trade-off between speed and overshoot introduced by the PD controller, whereas QSC's overshoot-free tracking reduces potential collisions with vascular walls. Moreover, the control input $u$ generated by the QSC approach, optimized using TD, smooths out the pulsating waves produced by the PD controller, avoiding potential hardware damage and ensuring refined control operations.

Further comparison is made between the cosine responses of the MSCR using two different controllers (see Movie \textcolor{blue}{S3}). As shown in \figref{fig:expresults}.(b), the QSC approach achieves fast and overshoot-free tracking of the cosine signal, exhibiting greatly improved control performance at peaks compared to the PD controller under the same parameter conditions. The estimated perturbations $\bar{x}_2$ demonstrate periodic behavior and gradual convergence, consistent with the previous theoretical analysis. Additionally, as shown in the performance error plots, the QSC approach effectively reduces the energy consumption of the controller, with a reported average steady-state error of $0.062$, outperforming the PD controller's value of $0.1364$. To validate the robustness of the controller in the presence of external disturbances (for clinical surgeries, consider vascular vibrations and hemodynamic impacts caused by the patient's cardiac pacing), a separate set of experimental tests was conducted in a wind disturbance environment starting from $15$ seconds (see Movie \textcolor{blue}{S4}). The results demonstrated that the control performance of the PD controller significantly deteriorated, with a steady-state error of $0.2005$. In contrast, the QSC approach maintained a high tracking accuracy, and the impact of high-frequency noise in the system output on the state estimate of the ESO is minor, with a reported steady-state error of $0.1299$.

Overall, the proposed quasi-static control framework effectively enhances the transient performance and steady-state error of the PD controller. However, the controller may encounter some challenges in higher-frequency signals (period $\leq 5$ $\mr{s}$), possibly due to the lag in the velocity control.
% 接着对比了MSCR采用两种控制器的余弦响应（见Movie S2）。如图\figref{fig:expresults}. (b)所示，QSC快速且无超调地跟踪上了余弦信号，并且在信号的峰值处的控制性能要明显优于相同参数条件下的PD控制器。$\bar{x}_2$预报的估计结果呈周期性并逐渐收敛，与(17)的理论分析结果相符。同时，QSC还有效地降低了控制器的能耗，平均稳态误差的报告值0.062要优于PD控制器的0.1364。

% 为了验证控制器在存在外部干扰（例如临床手术中病人心脏起搏引起的血管震动和血流冲击）下的鲁棒性，另一组对照试验从15秒开始暴露在风扰环境中（见Movie S3）。结果表明，PD控制器在存在外部扰动下的控制性能显著下降，稳态误差来到$0.2005$。相比之下，QSC依然保持较高的跟踪精度，且系统输出包含的高频噪声对ESO的估计结果的影响轻微，稳态误差报告为$0.1299$。

% 总体来说，提出的quasi-static control framework能够很好地提升PD控制器的瞬态性能和稳态误差。Algorithm 1对远端偏转角的高效量测得到充分验证。然而，控制器在高频正弦信号的情况下可能表现出一些挑战，这可能是由于机械臂速度控制的滞后引起，需要在未来进一步优化或采取额外的措施来提高性能。
% （对控制器参数的敏感性下降，k=2时PD发散而QSC不发散）

%  The efficient measurement of the distal angle using Algorithm \ref{algo:measureOfdistalangle} is thoroughly validated.

\section{Extension to Positional Control}
\label{sec:extension}
\begin{figure}[t]
    \centering
    \includegraphics[width=\columnwidth]{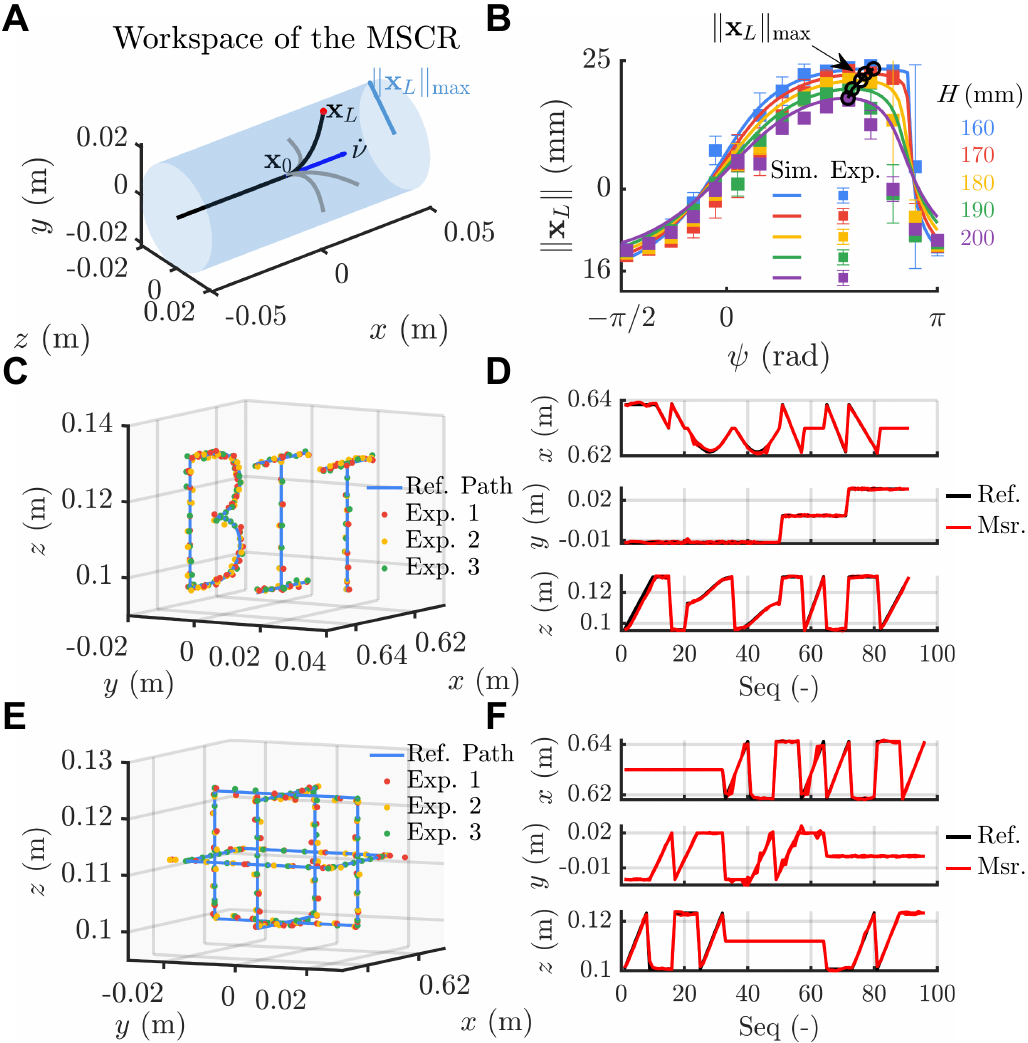}
    \caption{(\textbf{A}) Illustration of the workspace of the MSCR. (\textbf{B}) Comparison of displacement of the MSCR's distal end between simulation and experiment when the magnet is at different rotation angles and heights. (\textbf{C} and \textbf{E}) Path-following control of two paths with each of three experimental results. (\textbf{D} and \textbf{F}) Comparison of reference with average measured distal end position for each spatial coordinate.
    }
    \label{fig:pathfollowing}
\end{figure}

Some critical scenarios place control demands on the tip position of the MSCR, which usually involves computing the Jacobian matrix between the task-space and the actuation-space of the distal end position. The Jacobian we solved for the rotation angle can be conveniently applied to compute this matrix. As we mentioned previously, when the proximal end of the MSCR is fixed, the workspace of the tip position is limited and approximates a hemispherical surface. To expand the workspace, we add a linearly movable base to the MSCR, and its displacement is represented by $\nu$. The workspace of the movable MSCR is shaped like a cylinder in \figref{fig:pathfollowing}.\textbf{A}, and its radius is determined by the maximum displacement of the tip position $\|\mb{x}_L\|_{\max}$. \figref{fig:pathfollowing}.\textbf{B} compares the displacement of the MSCR's distal end between simulation and experiment when the magnet is at different rotation angles and heights. The simulation and experimental results show good consistency, indicating that the model can reasonably estimate the Jacobian matrix and maximum displacement (marked in black circles).

% 一些关键场景对MSCR的尖端位置提出控制需求，这通常涉及计算尖端位置的任务空间与驱动空间的雅可比矩阵。我们求解的关于旋转角度的雅可比可以很方便地被应用于计算这一矩阵。正如我们前面所提到的，当MSCR的proximal端固定时，尖端位置的工作空间是有限的，近似为半球面。为了拓展工作空间，我们为MSCR添加了可直线移动的底座，其位移量用$\nu$表示。可移动MSCR的工作空间形如图9.A中的圆柱体，其半径由尖端位置的最大位移量$\|\mb{x}_L\|_{\max}$决定。图9.B对比了外部磁铁在不同高度下的旋转角度与MSCR的最大位移量的关系，仿真和实验结果展现出良好的一致性，表明模型具有精确计算所需的Jacobian矩阵的潜力。

Let $\mb{x}_0(\nu)$ be the proximal end of the movable MSCR, and take the roll angle $\varphi=0$ as an example, the body coordinates satisfy the following kinematic equations:

% 设可移动MSCR的近端为$\mb{x}_0(\nu)$，以滚转角$\varphi=0$为例，机体坐标满足如下运动学方程：
\begin{equation}
    \mb{x}(s) = \mb{x}_0 + \left(\int_0^{s}\cos\theta(\eta)\mrd\eta\hb{x}_{\mc{G}}+\int_0^s\sin\theta(\eta)\mrd\eta\hb{y}_{\mc{G}}\right).
\end{equation}
The differential kinematics is then derived as
\begin{equation}
	\dot{\mb{x}}_L = \left[
	\frac{\pt\mb{x}_L}{\pt\nu}, \frac{\pt\mb{x}_L}{\pt\psi}
	\right]\begin{bmatrix}
		\dot{\nu} \\ \dot{\psi}
	\end{bmatrix}\triangleq\mr{B}\mb{u},
\end{equation}
where the components of the Jacobian matrix ($\mr{B}$) is given by
\[
    \frac{\partial \mb{x}_L}{\partial \psi}=-\int_0^L\sin\theta\frac{\pt\theta}{\pt\psi}(\eta)\mrd\eta\hb{x}_{\mc{G}}+\int_0^L\cos\theta\frac{\pt\theta}{\pt\psi}(\eta)\mrd\eta\hb{y}_{\mc{G}},
\]
and $\frac{\pt\mb{x}_L}{\pt\nu}=\begin{bmatrix}
        1, 0,  0
    \end{bmatrix}^\top$. The corresponding closed-loop controller can be designed as
\begin{equation}
    \mb{u} = \dot{\mb{x}}_L^r + k_{\mb{x}}\mr{B}^{\dagger}(\mb{x}_L^r - \mb{x}_L),
\end{equation}
in which $k_{\mb{x}}$ is a positive constant and $(\cdot)^{\dagger}$ denotes in the Moore-Penrose pseudo-inverse.

In the predicted MSCR workspace, two path-following control (PFC) tasks are designed to examine the controller's performance. The material properties of MSCR \#2 in Tab. \ref{tab:parameterOfMSCR} is employed, and the controller gain is set to $k_{\mb{x}}=0.5$. The distal end is marked, and a stereo camera captures the position. Each path was sampled with more than $90$ points. A path point is considered reached when the relative rate of change in positional error falls below $1$\%. Three sets of positional experiments were conducted for each path. The results indicate that the positional RMSE for the path in \figref{fig:pathfollowing}.\textbf{C} was $0.965\pm0.156\ (\mathrm{mm})$, while for \figref{fig:pathfollowing}.\textbf{E}, the positional error was $0.972\pm0.396\ (\mathrm{mm})$. The comparisons between the reference and the average measured distal end position for each spatial coordinate are shown in \figref{fig:pathfollowing}.\textbf{D} and \figref{fig:pathfollowing}.\textbf{F}, respectively. Overall, the positional error accounts for only $3.22$\% and $3.24$\% of the MSCR length, which demonstrates the effectiveness of our designed controller in achieving high-precision control over the MSCR's distal end.

% 两组路径被分别均匀采样了90个路径点，当误差的相对变化率小于1%时，视为达到了当前路径点。对每条路径都进行了三组跟踪实验，结果表明，\figref{fig:pathfollowing}.\textbf{C}中的路径的跟踪误差为$1.178\pm0.164\mathrm{mm}$，\figref{fig:pathfollowing}.\textbf{D}的跟踪误差为$1.245\pm0.552\mathrm{mm}$. 总的来说，位置误差仅占MSCR长度的3.92\%和4.15\%，这充分说明了我们设计的控制器在对MSCR的远端位置施行高精度控制的能力。

% Error 
% 在预测的MSCR工作空间内，设计了几组路径跟随任务来检验控制器对MSCR远端的控制性能。
% The maximum displacements are marked in black circles,
\section{Conclusions}
\label{sec:conclusions}
% 未来考虑移动磁铁使其能够获得更大的偏转

% 视觉算法
In this study, we achieved closed-loop deflection control of MSCRs by utilizing a single rotatable permanent magnet as the end-effector of the robot arm. The proposed kinematic model effectively characterizes the quasi-static states of MSCR in different configurations in a non-uniform magnetic field. The existence and uniqueness of Jacobian are supported theoretically, and the estimation of singularities is highly consistent with experimental results. The designed damping structure for Jacobian effectively mitigates the control law oscillations when crossing singularities. Experimental results demonstrated that the proposed QSC framework outperformed the PD controller in terms of steady-state accuracy and transient performance during trajectory tracking. Extensions are made for the path-following control of the distal end position, and the positional error is only sub-millimeter. This framework holds promise for precise intravascular navigation of MSCRs in the future. However, some unresolved issues remain, such as the inapplicability of camera-based visual platforms for in vivo surgeries. Future work will further explore the closed-loop control of MSCR combined with angiographic image guidance.

% Future work will focus on integrating the feeder device to enable MSCRs to perform more complex three-dimensional motions.

% 在这篇文章中，我们使用附着在机械臂末端的可旋转单一永磁铁实现了对MSCR的偏转控制。所提出的运动学模型很好地刻画了MSCR在非均匀磁场中不同配置下的quasi-static states. Jacobian的存在性和唯一性得到定理的支撑，对奇异值的估计与实验结果呈现高度一致。设计的Jacobian的阻尼结构有效缓解了在奇异值附近导致的控制律震颤。实验结果表明，提出的QSC在轨迹跟踪的稳态精度和瞬态性能上都要优于PD控制器。这一框架在未来可能能够被用于MSCR在血管内的精确导航。然而，仍有一些待解决的问题，例如camera-based的视觉平台在体内手术不再适用。未来工作将会聚焦于配合递送装置驱使MSCR实现更复杂的三维运动。

% 我们针对Jacobian的分析和求解还帮助应用于设计远端位置的控制器，在两组路径跟随实验中均取得了低的位置误差。

\appendix

\subsection{Proof of Theorem \ref{the:extTheta}}
\label{append:proof1}
\begin{proof}
It can be readily observed that the function $\sigma(s,\theta)$ is continuous on $\mc{R}:0\leq s\leq L,\theta^2+\theta^{\prime^2}\leq\infty$.
We show the derivative of $\sigma(s,\theta)$ with respect to $\theta$, written in the shorthand form as 
\begin{equation}
    \begin{aligned}
        \frac{EI}{A}\frac{\pt\sigma(s,\theta)}{\pt\theta}&=-\frac{\pt^2}{\pt\theta^2}(\mr{R}\mb{m}\cdot\mb{b})\\
        &=\mr{R}\mb{m}\cdot\left(\mb{b}+\nabla\mb{b}\mb{x}-\nabla^2\mb{b}\frac{\pt\mb{x}}{\pt\theta}\frac{\pt\mb{x}}{\pt\theta}\right)\\
        &+2\mb{e}_z\cdot\left(\mr{R}\mb{m}\times\nabla\mb{b}\frac{\pt\mb{x}}{\pt\theta}\right),
    \end{aligned}
\end{equation}
in which $\nabla^2\mb{b}$ is a 3-order tensor. Since $\forall s\in[0,L],\|\mb{p}_{\mc{A}}-\mb{x}(s)\|>0$, it follows that both the magnitude of $\mb{b}$ and of its higher-order gradient tensors are bounded. Using $\|\cdot\|$ for all tensors, we obtain that 
$\frac{\pt\sigma(s,\theta)}{\pt\theta}$ is uniformly bounded in $\mc{R}$ by
\begin{equation}
\begin{aligned}
        &\left|\frac{\pt\sigma(s,\theta)}{\pt\theta}\right|\\
        &<\frac{MA}{EI}\left[B+\|\nabla\mb{b}\|\left(\|\mb{x}\|+2\left\|\frac{\pt\mb{x}}{\pt\theta}\right\|\right)+\|\nabla^2\mb{b}\|\left\|\frac{\pt\mb{x}}{\pt\mb{\theta}}\right\|^2\right]\\
        &\leq\frac{MA}{EI}\left(B+3\|\nabla\mb{b}\|L+\|\nabla^2\mb{b}\|L^2\right)\triangleq K
\end{aligned}
\end{equation}
where we used the Cauchy-Schwarz inequality:
\begin{equation}
\begin{aligned}
    \left\|\frac{\pt\mb{x}}{\pt\theta}\right\|&=\sqrt{\left(-\int_0^s\sin\theta(\eta)\mrd\eta\right)^2+\left(\int_0^s \cos\theta(\eta)\mrd\eta\right)^2}\\
    &\leq\sqrt{s\int_0^s \sin^2\theta(\eta)+\cos^2\theta(\eta)\mrd\eta}\\
    &=s\leq L.
\end{aligned}
\end{equation}
 Hence, a Lipschitz constant can be taken as $K$, which depends only on the physical parameters of the MSCR and magnet. Given the fact that $\frac{\pt\sigma(s,\theta)}{\pt\theta^\prime}=0$, both $\tfrac{\pt\sigma(s,\theta)}{\pt\theta}$ and $\tfrac{\pt\sigma(s,\theta)}{\pt\theta^\prime}$ satisfy the Lipschitz condition. This conclusion, according to the existence theorems for solutions of BVP (\cite{keller2018numerical}), completes the proof. \qedhere
\end{proof}
\subsection{Proof of Theorem \ref{the:existenceOfSL-BVP}}
\label{append:proof2}
\begin{proof}
Define two unique functions $J^{(1)}(s)$ and $J^{(2)}(s)$ on $[0,L]$ as solutions of the respective initial value problems: (a) $LJ^{(1)}=\frac{\pt\sigma}{\pt\psi};\ J^{(1)}(0)=0,J^{(1)^\prime}(0)=0$ and (b) $LJ^{(2)}=0;\ J^{(2)}(0)=0,J^{(2)^\prime}(0)=1.$ The function $J(s)$ defined by $J(s)=J(s;v)\triangleq J^{(1)}(s)+vJ^{(2)}(s),0\leq s\leq L$ satisfies $J(0)=0$ and will thus be a solution of problem \eqref{eqn:SLprob} if $v$ is chosen such that $\phi(v)\triangleq J^{\prime}(L;v)=0$. The equation is linear in $v$ and has the single root $v=-J^{(1)^\prime}(L)/J^{(2)^\prime}(L)$, provided that $J^{(2)^\prime}(L)\neq 0$. 

In the proof of Theorem \ref{the:extTheta}, we conclude that $\frac{\partial\sigma}{\partial\psi}$ is bounded by $K$ when Assumption \ref{ass:1} holds. Consequently, invoking the comparison principle, we can assert that $\underline{\mc{J}^\prime}(s)<J^{(2)^\prime}(s)<\overline{\mc{J}^\prime}(s)$ for $s\in[0,L]$, where $\underline{\mc{J}}(s)$ and $\overline{\mc{J}}(s)$ are solutions to the following differential equations: $\frac{\mathrm{d}^2\underline{\mc{J}}(s)}{\mathrm{d} s^2}=-K\underline{\mc{J}}(s)$ and $\frac{\mathrm{d}^2\overline{\mc{J}}(s)}{\mathrm{d} s^2}=K\overline{\mc{J}}(s)$, subject to the same initial values as $J^{(2)}(s)$. Analytical solutions for $\underline{\mc{J}^\prime}(s)$ and $\overline{\mc{J}^\prime}(s)$ are found in $\underline{\mc{J}^\prime}(s)=\cos(\sqrt{K}s)$ and $\overline{\mc{J}^\prime}(s)=\cosh(\sqrt{K}s)$, respectively. Since the Lipschitz constant $K\in\mc{K}$, it is guaranteed that $\overline{\mc{J}^\prime}(L)=\cosh(\sqrt{K}L)>0$ and $\underline{\mc{J}^\prime}(L)=\cos(\sqrt{K}L)>0$, which leads to $J^{(2)^\prime}(L)>0$. Hence, $LJ^{(2)}$ has a nontrivial solution, and the SL-BVP \eqref{eqn:SLprob} has a unique solution $J(s)$.\qedhere
\end{proof}

\subsection{Proof of Theorem \ref{the:controllability}}
\label{append:proof3}
\begin{proof}
We first prove the existence of $\Theta$. The parameter $\psi$ can be separated from functions $\sigma(s,\theta;\psi)$ and $\frac{\pt\sigma(s,\theta;\psi)}{\pt\theta}$ as $\sigma(s,\theta;\psi)=\varsigma(s,\theta)\hb{m}_{\mc{G}}(\psi)$ and $\frac{\pt\sigma(s,\theta;\psi)}{\pt\theta}=\frac{\pt\varsigma(s,\theta)}{\pt\theta}\hb{m}_{\mc{G}}(\psi)$, where $\hb{m}_{\mc{G}}(\psi):\mathbb{R}\to\mathbb{R}^3$ is a smooth unit vector function. Thus, $\sigma(s,\theta;\psi)$ and $\frac{\pt\sigma(s,\theta;\psi)}{\pt\theta}$ are continuously differentiable in $\psi$. Note that $J_\psi=J(s;\psi)|_{s=L}$, where $J(s;\psi)=J^{(1)}(s;\psi)+vJ^{(2)}(s;\psi)$ is the linear composition of solutions to two initial-value problems. Based on the continuity theorem of the solutions with respect to parameters, the Jacobian $J_{\psi}$ is continuous over any closed interval $\Psi$. Consequently, $\vartheta(\psi)$ exhibits the same continuity property as $J_{\psi}=\frac{\mrd\vartheta(\psi)}{\mrd\psi}$. Then, as a function continuous on a closed interval is bounded there, it is found a value domain of $\theta_L$ by $\Theta\triangleq[\theta_L^\vee,\theta_L^\wedge]$. 

We then prove the controllability. Given an initial state $\theta_L(t_0)\in\Theta$, the corresponding rotation angle $\psi(t_0)=\vartheta^{-1}(\theta_L(t_0))\in\Psi$. Then, for any final state $\theta_L(t_f)\in\Theta$, there always exists a corresponding rotation angle $\psi(t_f)=\vartheta^{-1}(\theta_L(t_f))$ by the intermediate value theorem. Hence, the admissible control input can be taken as $u=\frac{\psi(t_f)-\psi(t_0)}{t_f-t_0}$, and thus the controllability is proved. Throughout the proof, Assumption \ref{ass:1} guarantees the existence of $\theta(s)$, and Assumption \ref{ass:2} guarantees the existence and uniqueness of $J(s)$. \qedhere
\end{proof}

% \addtolength{\textheight}{-3cm}   % This command serves to balance the column lengths

%%%%%%%%%%%%%%%%%%%%%%%%%%%%%%%%%%%%%%%%%%%%%%%%%%%%%%%%%%%%%%%%%%%%%%%%%%%%%%%%

%%%%%%%%%%%%%%%%%%%%%%%%%%%%%%%%%%%%%%%%%%%%%%%%%%%%%%%%%%%%%%%%%%%%%%%%%%%%%%%%

%%%%%%%%%%%%%%%%%%%%%%%%%%%%%%%%%%%%%%%%%%%%%%%%%%%%%%%%%%%%%%%%%%%%%%%%%%%%%%%%

\section*{ACKNOWLEDGMENT}

The author would like to thank Siyi Wei and Zhanxin Geng for their invaluable assistance and contributions to this work. Additionally, the author extends gratitude to all the reviewers for their significant contributions to improving this paper.

%%%%%%%%%%%%%%%%%%%%%%%%%%%%%%%%%%%%%%%%%%%%%%%%%%%%%%%%%%%%%%%%%%%%%%%%%%%%%%%%
\bibliographystyle{IEEEtran}
\bibliography{references.bib}
\vspace{-1cm}
\begin{IEEEbiography}[{\includegraphics[width=1in,height=1.25in,clip,keepaspectratio]{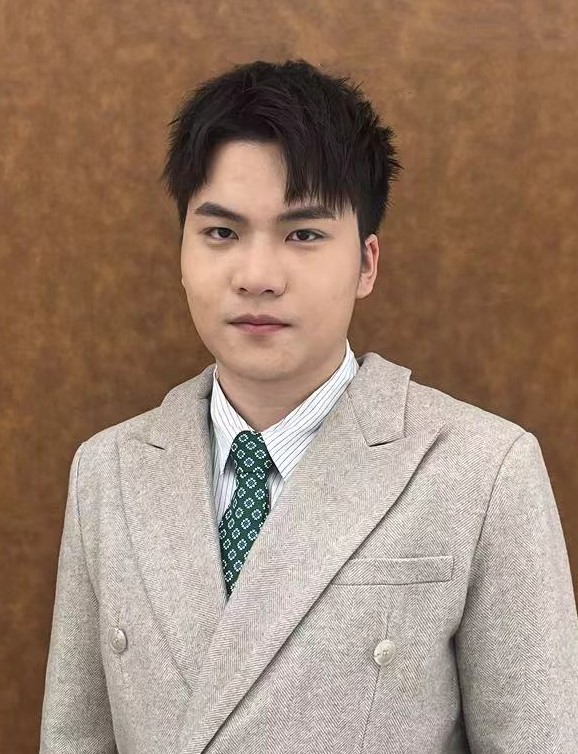}}]{Zhiwei Wu} was born in Fujian, China, in 2000. He received a B.E. degree from the College of Information Science and Technology, Beijing University of Chemical Technology, Beijing. He is currently working towards a Ph.D. degree in the Department of Automation at the Beijing Institute of Technology, Beijing. His research interests include surgical robotic systems and the control of magnetic soft continuum robots.
\end{IEEEbiography}

\vspace{-1cm}
\begin{IEEEbiography}
[{\includegraphics[width=1in,height=1.25in,clip,keepaspectratio]{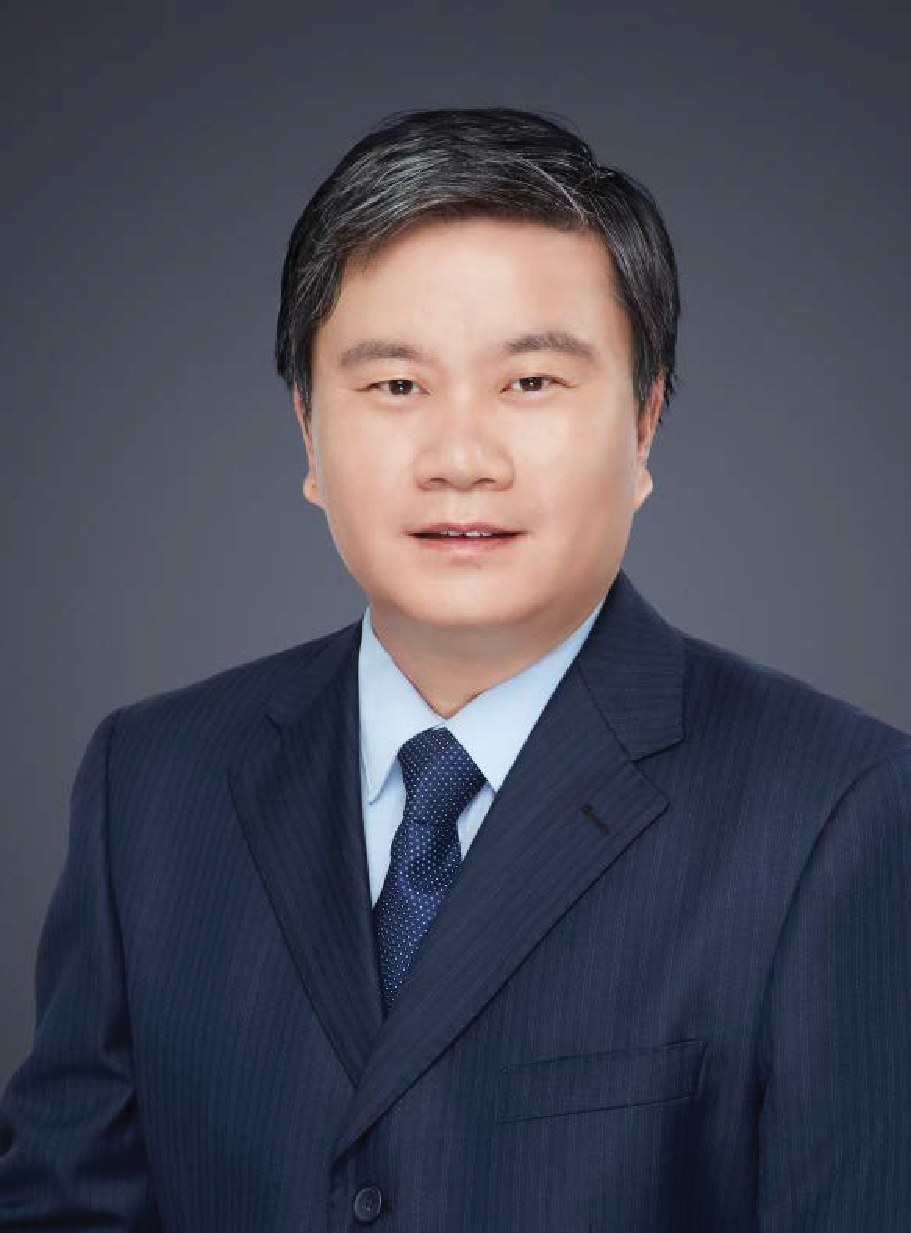}}]{Jinhui Zhang} received the Ph.D. degree in control science and engineering from the Beijing Institute of Technology, Beijing, China, in 2011. He was an Associate Professor with the Beijing University of Chemical Technology, Beijing, from 2011 to 2016. He joined the Beijing Institute of Technology in 2016, where he is currently a Professor. His research interests include networked control systems and composite disturbance rejection control.
\end{IEEEbiography}

\end{document}